\pdfoutput=1

\documentclass[11pt]{article}

\usepackage[]{acl}

\usepackage{times}
\usepackage{latexsym}
\usepackage{amsmath}

\usepackage{booktabs}
\usepackage{graphicx}
\usepackage{todonotes}

\usepackage{tikz}
\def\checkmark{\tikz\fill[scale=0.4](0,.35) -- (.25,0) -- (1,.7) -- (.25,.15) -- cycle;} 

\usepackage{adjustbox}
\usepackage{hyperref}

\usepackage[T1]{fontenc}

\usepackage[utf8]{inputenc}

\usepackage{microtype}

\usepackage{lipsum}

\newcommand{\newterm}[1]{{\bf #1}}
\newcommand{\KE}{K{\small NOWLEDGE}E{\small DITOR }}
\newcommand{\lmkb}{\texttt{LMs-as-KBs} }
\newcommand{\AP}{A{\small UTO}P{\small ROMPT }}

\title{A Review on Language Models as Knowledge Bases}

\author{Badr AlKhamissi\thanks{\;\;Equal Contribution} \And Millicent Li\footnotemark[1] \AND Asli Celikyilmaz\thanks{\;\;Equal Supervision} \And Mona Diab\footnotemark[2] \\ \\ Meta AI \\ \texttt{\{bkhmsi, millicentli, aslic, mdiab, ghazvini\}@fb.com}  \And Marjan Ghazvininejad\footnotemark[2]  }

\begin{document}
\maketitle
\begin{abstract}
    
   
    Recently, there has been a surge of interest in the NLP community on the use of pretrained Language Models (LMs) as Knowledge Bases (KBs). Researchers have shown that LMs trained on a sufficiently large (web) corpus will encode a significant amount of knowledge implicitly in its parameters. The resulting LM can be probed for different kinds of knowledge and thus acting as a KB. This has a major advantage over traditional KBs in that this method requires no human supervision. In this paper, we present a set of aspects that we deem an LM should have to fully act as a KB, and review the recent literature with respect to those aspects.\footnote{For an updated paper-list please check our website: \href{https://bkhmsi.github.io/lms-as-kbs/}{https://bkhmsi.github.io/lms-as-kbs/}.}

\end{abstract}



    
            
            
    
    

\section{Introduction}


The impact of Pretrained Language Models (\newterm{LMs}) on Natural Language Processing (\newterm{NLP}) research can be described as nothing short of transformative. It has moved the field from \textit{feature engineering}  \cite{och-etal-2004-smorgasbord, zhang-nivre-2011-transition} and \textit{architecture engineering} \cite{Chung2014EmpiricalEO, kim-2014-convolutional, Bahdanau2015NeuralMT, vaswani17attention} to the \textit{pre-train and fine-tune} paradigm \cite{Radford2018ImprovingLU, dong_NEURIPS2019_c20bb2d9, lewis_retrieval-augmented_2021}, and lately the \textit{pre-train, prompt, and predict} paradigm \cite{Liu2021PretrainPA}. LMs pretrained on a large corpus of web data have been shown to contain different kinds of knowledge implicitly in their parameters without the need for any human supervision. This includes: world knowledge \cite{petroni_language_2019, rogers_primer_2020}, relational knowledge \cite{safavi2021relational}, commonsense knowledge \cite{Da2021AnalyzingCE}, linguistic knowledge \cite{Peters2018DissectingCW, Goldberg2019AssessingBS, tenney2018what}, actionable knowledge \cite{Huang2022LanguageMA} and more. This access to knowledge is crucial for LMs to achieve state-of-the-art results on various downstream tasks. However, as is the case with most neural systems, knowledge in LMs is encoded in a diffused manner, making it generally difficult to interpret and hard to update.

\begin{figure}
    \centering
    \includegraphics[width=1\linewidth]{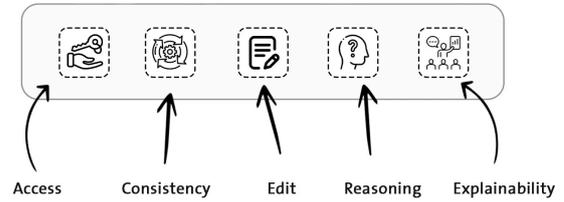}
    \caption{Aspects of \lmkb as presented in this work}
    \label{fig:lmkb}
\end{figure}

Despite these recent breakthroughs, we often do not have full control over the behavior of LMs. As a result, utilizing these models in real-world scenarios is often unsuccessful. On the other hand, Knowledge Bases (\newterm{KB}) are easier to control. Here, KBs refers to a data structure that stores relational information in the form of triples connecting two triplets of entities by symbolic relations (e.g. $\langle$ \texttt{Cairo}, \texttt{CapitalOf}, \texttt{Egypt} $\rangle$). They often follow rule-based heuristics, rendering them predictable, in addition to possessing large knowledge coverage, which primes them for use in real-world systems. These models are often used as chatbots and virtual assistants, where controlled generation of output is necessary to ensure appropriate responses \cite{chen-dialogue-2017}. Therefore, KBs are a natural solution to access specific gold-standard relational information. They are repositories of knowledge, for both structured and unstructured data, and can be seamlessly queried and updated by an end user. 


Since KBs can access and update relational knowledge easier than LMs can, one question naturally arises: how can we control the repository of knowledge stored implicitly in the weights of a LM as similarly as KBs can? This question was first introduced in seminal work by \citet{petroni_language_2019} and has since ignited the interest of the community with the goal of instilling LMs with desirable properties of KBs.

Several works have already approached improving LMs through the lens of KBs: \citet{petroni_language_2019, dhingra_time-aware_2021, wang_language_2020, heinzerling_language_2021, sung2021language}. Many of these works include updating factoids stored within the parameters of LMs \cite{de_cao_editing_2021, mitchell_fast_2021, hase_language_2021} to creating new methods for extracting factual knowledge \cite{petroni_language_2019}. Despite significant progress towards achieving parity between LMs and KBs, LMs still lack specific aspects that KBs have. For example, given the cloze phrases ``\texttt{Albert Einstein was born in [MASK]}'' and ``\texttt{The hometown of Albert Einstein is [MASK]}'', a user of a KB can map both queries to one triplet $\langle$ \texttt{Albert Einstein}, \texttt{BornIn}, \textbf{X} $\rangle$ that the KB readily understands and thus consistently returns the same city. On the other hand, LMs may not be as consistent, yielding potentially different answers for the same underlying factual questions. 

In this survey, we propose to consolidate the work on \texttt{LMs-as-KBs} within one cohesive framework, with a focus on aspects related to KBs that we think are useful to integrate into LMs. Few survey papers exist that evaluate LMs in this context. For instance, \citet{wei2021knowledge} evaluate knowledge-enhanced pretrained LMs by delineating the types of knowledge that can be integrated into existing LMs. \citet{safavi2021relational} divide relevant work according to the level of supervision provided to the LM by a KB. Similarly, \citet{colonhernandez2021combining} cover the integration of structural knowledge into LMs but forgo implicit knowledge. Our study of \lmkb from our perspective is unique compared to the focus of existing survey papers. We aim to present the current landscape of \lmkb research and highlight the existing challenges that LMs face when applied in practice. 


We observe the recent advances in LMs and explore them with respect to aspects that we find necessary for LMs to become as functional and utilizeable as KBs: \textit{access}, \textit{consistency}, \textit{editability}, \textit{reasoning}, and \textit{explainability}. We further highlight where we are now in terms of the \lmkb framework as well as potential work for the future. Finally, we discuss the remaining challenges in the full adoption of \lmkb and propose directions for future research. This survey is structured as follows:

\begin{enumerate}
    \item First, an overview of KBs and LMs and their intersection for \lmkb
    \item Second, the enumeration of the aspects of \lmkb
    \item Third, a brief summarization of each aspect with respect to recent work
\end{enumerate}


    


We hope that by highlighting the aspects of \lmkb, we can consolidate knowledge in this ever-growing field of research. We envision that our work can provide a path for those new to the area of research to better improve LMs to be just as good, if not eventually better, than KBs.



\section{Preliminaries}
In this section, we define what LMs and KBs are, characterize the functions and attributes of KBs, and make direct comparisons between both LMs and KBs to highlight the current limitations of LMs in the context of the \lmkb framework.

\subsection{Knowledge Bases}
\label{sec:kbs}
KBs usually adhere to a manually engineered schema that dictates the possible set of entities and relations and the interactions between them. Such a rigid schema facilitates different kinds of complex operations over the data (e.g. multi-hop reasoning) and ensures accurate, consistent and explainable answers. Examples of KBs include Wikidata \cite{wikidata} and the ATOMIC \cite{sap2019atomic} (See Appendix \ref{app:kgs} for further details).

\subsection{Language Models}
In the context of this paper, we use the term ``LM" to refer to a deep neural LM that is pre-trained on a large amount of unlabeled text in a self-supervised setting such as masked language modeling. Examples include transformer-based models \cite{vaswani17attention} such as BERT \cite{devlin-etal-2019-bert}, GPT-3 \cite{gpt3}, BART \cite{lewis-etal-2020-bart} and T5 \cite{Raffel2020ExploringTL}.


\subsection{LMs-as-KBs}

KBs lack the flexibility that LMs offer in terms of extendability and expressability. KBs also require significant human effort to build and maintain. For example, populating KBs involves extracting huge amounts of relational data from unstructured text by using complex NLP pipelines such as entity linking and coreference resolution \cite{petroni_language_2019}. In contrast, LMs are able to implicitly capture this information without any supervision. 

To better frame the current limitations of LMs with respect to the capabilities of KBs, we observe a total of five aspects of KBs that we would like LMs to excel at in order to be considered a KB. We consider \textbf{Access}, \textbf{Edit}, \textbf{Consistency}, \textbf{Reasoning}, and \textbf{Explainability and Interpretability}. The latter three are not implicitly done, but are easier to ensure within KBs than LMs.




\paragraph{Access}
KBs are often simple to access via manual querying of terms using specific querying languages. On the other hand, LMs cannot be queried explicitly as the knowledge is not directly encoded in specific locations in the model. Current research has focused on \textit{how} we can query LMs similarly to how we can query KBs, focusing on specific access patterns for different types of knowledge. For example, knowledge encoded in LMs can be accessed through probing via fill-in-the-blank (cloze) prompts \cite{taylor1953cloze} and through traditional downstream finetuning. However, there is still more work needed to improve LMs for efficient and direct access.

\paragraph{Edit}
Since LMs are pre-trained on a certain snapshot of data from a specific time, the knowledge it learns can be outdated (e.g. the population of some country), incorrect \cite{Lin2021TruthfulQAMH}, biased or toxic \cite{Gehman2020RealToxicityPromptsEN, bender21}. Perhaps more importantly, in the context of privacy, LMs may memorize sensitive information that needs to be removed \cite{carlini_extracting_2021}. Further, new information is created all the time. For example, most available LMs today would have never seen information related to the COVID-19 Omicron variant. However, updating a specific fact in a LM is not straightforward, as facts are encoded in the weights of the model in a distributed fashion, making them inaccessible or uninterpretable \cite{mitchell_fast_2021}. The naive way of re-training the whole model on an updated set is expensive, especially with the ever-increasing size of current LMs \cite{gpt3}. There are increasing evidence that show that scaling LMs to larger sizes is not the solution to generating factually correct information \cite{lazaridou_mind_2021, Gehman2020RealToxicityPromptsEN, Lin2021TruthfulQAMH}. As a result, this would also result in catastrophic forgetting \cite{wallat_bertnesia_2021}. Changing a single weight may have a ripple effect that affects a large number of other implicitly memorized facts. Therefore, this task of knowledge editing is of utmost importance, especially when considering LMs in the context of KBs.

\paragraph{Consistency}
Language is multifaceted: the same meaning can be expressed in multiple forms. Structured KBs are built with consistency in mind; several efficient algorithms have been proposed to detect inconsistencies in KBs \cite{andersen01} so such conflicts can be easily resolved, while other work aimed at quantifying the degree of which inconsistency arises in KBs \cite{muino11}. Therefore, in the face of such language variability, we should expect LMs to behave consistently under semantically equivalent contexts, even across different languages. Note that consistency does not imply correctness, as a KB can be consistent but store factually incorrect information. Similarly, a LM can have an incorrect belief about a certain fact but that belief is consistent across different queries. 

\paragraph{Reasoning}

In a KB, it can be simple to follow the path of reasoning. For instance, given the KB triplet $\langle$ \texttt{Cairo}, \texttt{CapitalOf}, \texttt{Egypt} $\rangle$, the KB can sensibly reason that \texttt{Cairo} is in \texttt{Africa} given that it has another relation explicitly stating that \texttt{Egypt} is a part of \texttt{Africa}. On the other hand, recent work has shown that LMs can perform different forms of reasoning when finetuned on datasets that elicit reasoning capabilities within LMs \cite{west_symbolic_2021, talmor_leap--thought_2020, talmor_olmpics_2020, hase_language_2021}. Reasoning is not readily obvious and difficult to facilitate, as LMs have been shown to perform poorly on some types of reasoning such as structured reasoning \cite{kassner2020negated}.

\paragraph{Explainability and Interpretability}
Given a KB triplet, nodes and links can easily be identified, and the answer can easily be inferred. However, in LMs, knowledge is rarely understood by simply looking at the output. Moreover, the location of the parameters in which the knowledge comes from is unknown. Under perfect circumstances, one would want LMs to be explainable and interpretable to the end user, especially for stakeholders with no prior understanding of NLP \cite{Lakkaraju2022RethinkingEA}. However, current research on LMs is far from achieving this goal, as many of the newer techniques focus on black box rather than white box techniques \cite{danilevsky_survey_2020}. As a result, imbuing explainability and interpretability in LMs is core to improve \lmkb. We make the distinction, however, that KBs are inherently explainable. On the other hand, interpretability, while not latent in KBs, is still an important aspect to apply to LMs \cite{Lipton2018TheMO} that we also focus on for the purposes of the survey.

\begin{figure*}
    \centering
    \includegraphics[width=1\linewidth]{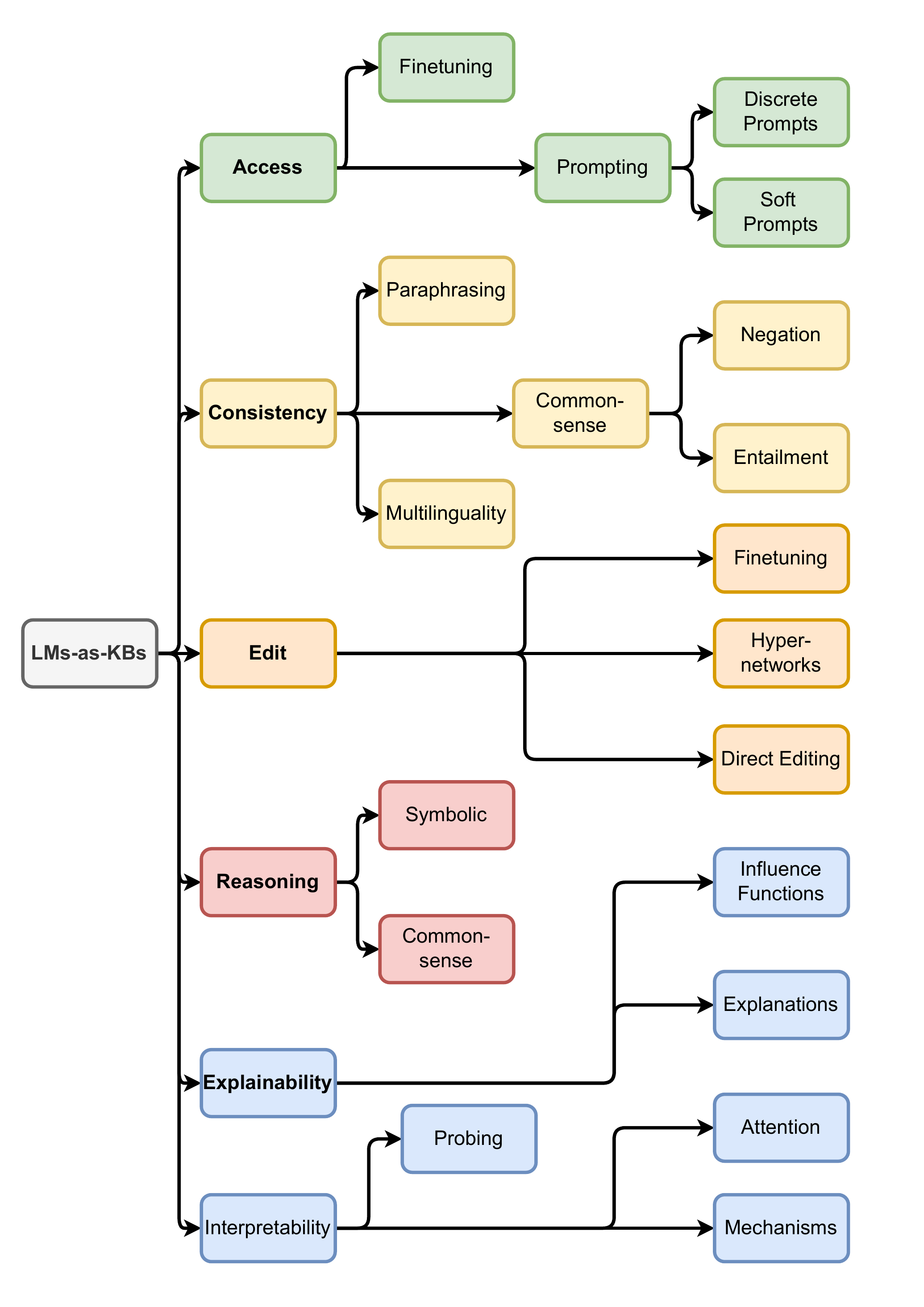}
    \caption{Finegrained Aspects of \lmkb}
    \label{fig:taxonomy}
\end{figure*}

\section{Accessing Knowledge}






KBs can simply be accessed by querying for specific entity nodes given the starting node and edge corresponding to the relation. In contrast, it is more difficult for LMs to access for specific pieces of knowledge. However, previous research has shown that LMs have the ability to be efficient few-shot and zero-shot learners, which shows that knowledge learned during pretraining can be potentially accessible by finetuning or prompting \cite{brown2020language}. Specifically, finetuning LMs on downstream tasks has shown to be an effective method to tune and elicit specific knowledge for evaluation  \cite{Radford2018ImprovingLU, dong_NEURIPS2019_c20bb2d9}. On the other hand, \citet{Liu2021PretrainPA} has recently shown that prompting shows promise as an effective method to directly access this knowledge without any additional finetuning. 





\subsection{Finetuning}

The knowledge stored in a LM is most often inaccessible to the end-user as compared to a KB. Therefore, to retrieve specific pieces of information from a LM, one prevailing method is to finetune the model on a relevant downstream task (e.g. commonsense question answering) so it can make way for the required knowledge to surface in the output during evaluation. Previous work has shown that most knowledge encoded in a LM are acquired during pretraining, while finetuning just learns an interface to access that acquired knowledge \cite{Da2021AnalyzingCE, wallat_bertnesia_2021}.

\subsection{Prompting}



The ever-increasing size of LMs make them expensive to finetune and store in practice, despite some architectural innovations that overcomes some of these challenges \cite{Houlsby2019ParameterEfficientTL}. On the other hand, prompting has emerged as an alternative method to extract the wanted knowledge directly from a LM \cite{qin2021learning}. These prompts are often difficult to craft \cite{Adolphs2021HowTQ, qin2021learning, jiang_how_2020}, and small changes in prompts can result in large performance differences, which can in turn affect consistency (see Section \ref{sec:consistency}). Research following the \textit{pretrain, prompt, and predict} paradigm \cite{Liu2021PretrainPA} utilize prompts to induce more knowledge from LMs \cite{petroni_language_2019, davison-etal-2019-commonsense, jiang_how_2020}. Prompting can be divided up into several categories, including \textbf{Discrete Prompts} and \textbf{Soft Prompts}.


\paragraph{Discrete Prompts}
Prompting often enables LMs to learn a specific subtask without extensive finetuning. This paradigm gives the model a familiar query format, which in turn leads to better responses. Many papers tackle prompting from the view of cloze-style like in \citet{petroni_language_2019, davison-etal-2019-commonsense, jiang_how_2020, talmor_olmpics_2020, dhingra_time-aware_2021, liu2021probing} in which these works use prompting to extract specific knowledge, such as commonsense \cite{davison-etal-2019-commonsense}, temporal \cite{dhingra_time-aware_2021, liu2021probing} and factual \cite{petroni_language_2019}, in a format known as \textit{Discrete Prompts}. Prompting has also been applied to specific domains, such as biomedical, to extract domain-specific knowledge \cite{sung2021language}. It was first introduced in \citet{Radford2018ImprovingLU, Radford2019LanguageMA} where it was shown that it could achieve decent zero-shot performance by crafting the right prompt. Knowing that prompting worked well, the same knowledge was applied to smaller LMs. \citet{schick2021exploiting, schick2021its, gao2021making} find that prompting smaller LMs improves performance, especially over supervised baselines. Others also take advantage of this improved performance and evaluate other discrete prompting approaches, such as through entailment \cite{wang2021entailment} and label token optimization \cite{Zhang2021DifferentiablePM}.

The challenge of crafting the ideal prompt for specific task is a non-trivial one. \citet{shin2020autoprompt} tackles this by taking a gradient-based search to find the appropriate prompt for a specific task. \AP creates a template automatically to do so. On the other hand, \citet{LoganIV2021CuttingDO} takes a more manual approach to crafting prompts with comparable accuracy to manual prompts by creating \textit{null prompts}, those of which are task-agnostic and are a simple concatenation of the input and a \texttt{[MASK]} token.


\paragraph{Soft Prompts}
Other approaches fall under the categorization of \textit{Soft Prompts}. Those are prompts that are represented by continuous word vectors, used as input and tuned, while keeping the remainder of the model unchanged. \citet{li2021prefixtuning} achieve comparable results on generation while using very few of the model's actual parameters by proposing prefix-tuning; task-specific vectors that can be tuned. Meanwhile, specific to extracting factual knowledge, \citet{qin2021learning, zhong2021factual} find that \textit{Soft Prompts} carry an advantage over \textit{Discrete Prompts} since they are more expressive and can represent multiple contexts simultaneously. \citet{zhong2021factual} takes a gradient-based approach to soft prompting while \citet{qin2021learning} improves upon \AP via continuous word vectors. A number of papers show that soft prompting, when optimized for specific tasks such as relation extraction and natural language understanding, can achieve better performance with minimal tuning \cite{lester2021power, han2021ptr}.

Other methods have been used without editing model parameters. First introduced with GPT-3 \citet{brown2020language}, \textbf{in-context learning}, which adds extra information to the model in the form of in-context demonstrations can improve performance. However, these works are still far from achieving human-level performance \cite{gao2021making, liu2021makes}.

\section{Consistency}
\label{sec:consistency}


LMs are shown to suffer from a lack of consistency in their answers \cite{elazar_measuring_2021}. For example, they can provide different results when queried for the same fact but under a different wording (i.e. a paraphrase). In this section, we consider consistency in light of three different contexts: \textbf{Paraphrasing}, \textbf{Commonsense} and \textbf{Multilinguality}.

\paragraph{Paraphrase}
\citet{bhagat13} define the term \textit{quasi-paraphrases} as `sentences or phrases that convey approximately the same meaning using different words'. The word \textit{approximately} is key here, since it does not assume the strict and logical equivalence of paraphrase. This fuzzy definition allows for a broader set of samples to be considered as paraphrases, and is of interest when considering consistency for LMs. Therefore, one method for measuring the consistency of a model is to probe it using paraphrases of the same relation for a specific subject, and test whether it always predicts the same object or not \cite{kassner2020negated, Ettinger20, elazar_measuring_2021}. To that end, several benchmarks have been proposed to measure consistency of LMs \cite{ravichander-etal-2020-systematicity, elazar_measuring_2021}, while other datasets have been adapted for that purpose \cite{levy-etal-2017-zero, wang2021}. For instance, \citet{de_cao_editing_2021} and \citet{mitchell_fast_2021} employ back-translation to generate paraphrases \cite{sennrich-etal-2016-improving, wieting-gimpel-2018-paranmt} for measuring consistency after editing factual knowledge in LMs. To mitigate this lack of consistency, \citet{elazar_measuring_2021} include a new term in their loss function that aims to minimize the Kullback-Leibler (\newterm{KL}) divergence for instance between the output distribution of different paraphrases.


\paragraph{Commonsense}
Consistency, however, is not limited to paraphrasing. Previous work explores the brittleness of LMs and the addition of a negation element (e.g. not) to a probe \cite{kassner2020negated, Ettinger20}. Specifically, a LM can maintain two contradictory beliefs in its parameters, such as ``\texttt{Birds can fly}'' and ``\texttt{Birds cannot fly}'', showing that they are insensitive to the contextual impacts of negation. Further, \citet{kassner2020negated} show an equal effect when mispriming the probe with a misleading distractor (e.g. \texttt{Talk? Birds can [MASK]}). Robust LMs will not only be consistent under different paraphrases and negations, but also under entailment. \citet{hase_language_2021} measure consistency under entailment, including contrapositives, after updating beliefs of a LM. Specifically, they adapt the LeapOfThought dataset \cite{talmor_leap--thought_2020} such that each datapoint has a main fact (e.g. \texttt{A viper is a vertebrate}) that they update and an entailed claim (e.g. \texttt{A viper has a brain}) that they check the truth value of with respect to the model's updated beliefs. Similar to the efforts done for overcoming lack of consistency under different paraphrases, \citet{hase_language_2021} add another loss term to their objective function to minimize the error across entailed data. On the other hand, \citet{kassner_beliefbank_2021} use a feedback mechanism that issues relevant information from a symbolic memory of beliefs as input to the system during test-time in order to improve consistency under entailment.

\paragraph{Multilingual}
Crosslingual LMs that are trained on several languages, such as XGLM \cite{Lin2021FewshotLW}, must be consistent across the languages they support. For example, the same probe queried in different languages must provide the same fact. \citet{liu_knowledge_2021} design a knowledge-based multilingual LM pretraining framework using Wikidata \cite{wikidata} that shows improvement on crosslingual NLP tasks, but neither they nor other work measure consistency under multilinguality.
\section{Model Editing}

There have been a number of works that address the problem of model editing, with strategies ranging from simple finetuning to making use of an external memory for adding or replacing factual knowledge. To that end, \citet{de_cao_editing_2021} describe a set of rules that editing methods should conform to: \textbf{Generality} implies that the editing method should be capable of changing the facts of \emph{any} LM that is not specifically trained on adaptability. For example, training using meta-learning is one way to make the models editable using a few gradient steps \cite{Finn2017ModelAgnosticMF}. \textbf{Reliability} means that editing a LM should only affect the targeted fact while retaining other unrelated information. \textbf{Consistency} signifies that changes should be consistent across semantically equivalent inputs. \citet{hase_language_2021} measure consistency under entailment (i.e. changing one fact must change other entailed facts in the LM), which \citet{mitchell_fast_2021} call the equivalence neighborhood.   

\subsection{Task Formulation}

To put it formally, given a LM $f_{\theta}$ that contains a collection of implicit factual knowledge $\mathcal{F}: \{(x, y)_i\}$ in its parameters $\theta$ by mapping an input $x_i$ to a potentially undesired output $y_i$, the goal is to obtain a new parameter set $\theta^*$ that conforms to a dataset of revisions $\mathcal{D}: \{(x,\hat{y})_j\}$ by predicting the desired output $\hat{y}_j$ for $x_j$. In addition, the new model $f_{\theta^*}$ should alter the base model's output on the equivalence neighborhood $\mathcal{P}^{x}$ of inputs in $\mathcal{D}$ (i.e. related input/output pairs such as paraphrases of $x$) while leaving the model's behavior intact on other unrelated inputs $\mathcal{O}^{x}$.

\subsection{Finetuning}

\paragraph{Baseline Methods}
One natural strategy to solve this problem is to finetune the model in question or retrain it from scratch on a modified training corpora that is consistent with the new facts. However, such a method would be expensive and impractical for modifying a few datapoints, especially for large LMs. Another approach is to construct a collection of supporting evidences for the modified facts and use it to finetune the model by minimizing a per-instance loss. This method can achieve high performance on the modified facts $\mathcal{D}$, but can significantly degrade the model's performance on the unmodified facts \cite{zhu_modifying_2020}. To obtain a reasonable accuracy on both, one can include evidences from $\mathcal{D}$ and the set of facts that should not be modified $\mathcal{O}^{x}$ in every iteration while finetuning. However, as discussed below, further tricks need to be employed to retain the accuracy on the unmodified facts to avoid catastrophic forgetting. 

\paragraph{Constrained Finetuning} \citet{zhu_modifying_2020} tackle this problem by enforcing a norm-based constraint on the model's weights $\theta$ while finetuning the model on the dataset of revisions $\mathcal{D}$ such that it minimally interferes with the facts that should not be modified. However, such a constraint on the parameter space ignores the highly non-linear nature of LMs \cite{de_cao_editing_2021}. 

\subsection{Hyper Networks}

Another class of methods uses a set of small neural networks (also known as a hyper-networks or learned optimizers) that learns to predict the shift in weights for editing a targeted fact \cite{HyperNetworks}. Specifically, \citet{de_cao_editing_2021} propose \KE that additionally applies a constraint on the update in the function space as opposed to the parameter space like in \citet{zhu_modifying_2020}. The intuition behind this is to predict identical output distributions to the original one for all unrelated inputs $\mathcal{O}^x$. However, this method fails when editing very large models. Therefore, to scale to larger models, \citet{mitchell_fast_2021} propose Model Editing Networks with Gradient Decomposition (\newterm{MEND}) that trains hyper-networks on the standard finetuning gradient of a given correction. The trick they employ is decomposing the gradient into its rank-one outer product form to learn a function $g$ that scales nicely with a model's dimensionality, making it much less computationally expensive than prior methods. In a parallel line of work, \citet{hase_language_2021} introduce a training objective for Sequential, Local, and Generalizing model updates (\newterm{SLAG}). They show that SLAG outperforms previous methods when updating multiple beliefs in sequence.

\subsection{Direct Editing}

\citet{meng2022locating} propose another method they call Rank-One Model Editing (\newterm{ROME}) for modifying factual knowledge in large neural networks. Specifically, they consider the transformer MLP modules as simple key-value memories \cite{geva2021transformer}, and editing a specific fact is just a matter of locating the relevant MLP weights by causal tracing then writing to it directly the new key-value pair using a rank-one modification.

\section{Reasoning}

It has been a long-standing goal of AI to reason over explicit knowledge given to it in order to reach conclusions \cite{McCarthy_Programs59, NewellLogic56, Metaxiotis2002ExpertSI}. The advent of LMs has brought this dream closer to reality \cite{Clark2020TransformersAS}. Concretely, LMs are shown to leverage the knowledge they learn during pretraining to perform well on reasoning tasks expressed in natural language rather than in formal representations. Such tasks include: commonsense \cite{Da2021AnalyzingCE}, natural language inference \cite{bowman-etal-2015-large}, mathematical \cite{Saxton2019AnalysingMR, Polu2022FormalMS}, rule-based \cite{Clark2020TransformersAS}, inductive \cite{Misra2021OnSC} and abductive reasoning \cite{Bhagavatula2020AbductiveCR}. However, we focus on types of reasoning that KBs have been shown to perform better on NLP tasks.

\subsection{Symbolic Reasoning}

\paragraph{Logical Reasoning} 
Previous work show that LMs can perform rule-based reasoning by emulating the process of binding facts with low level first order logic rules to deduce a conclusion \cite{Clark2020TransformersAS, talmor_leap--thought_2020, Gontier2020MeasuringSG}. LMs can also generate proofs demonstrating their `chain of thought' \cite{Saha2020PRoverPG, tafjord-etal-2021-proofwriter, Polu2022FormalMS}. \citet{Wei2022ChainOT} show that inducing a prompt that mimics the reasoning process improves the performance on reasoning tasks such as math word problems.

\paragraph{Mathematical Reasoning}
In another line of work, \citet{Nye2021ShowYW} show that LMs can perform complex multi-step computations when asked to generate the results of intermediate steps. \citet{Polu2022FormalMS} show that by using expert iteration (i.e. proof search \cite{Polu2020GenerativeLM} interleaved with curriculum learning) a LM can solve complex mathematical problems.  

\paragraph{Limitations}
Despite these successes, the best models are still unable to chain more than 2 or 3 non-trivial steps of complex reasoning \cite{Polu2022FormalMS}. Further, it largely remains an open question of whether the LMs are indeed ``reasoning'' or just emulating the thought process of humans \cite{Wei2022ChainOT}. 

\subsection{Commonsense Reasoning}
Commonsense reasoning is the ability to reason about the underlying nature of situations humans encounter on a day-to-day basis such as the effects of Newtonian physics and the intentions of others. LMs are shown to possess a certain amount of commonsense knowledge in their parameters \cite{petroni_language_2019, davison-etal-2019-commonsense, Zhou_Zhang_Cui_Huang_2020, cui-etal-2021-commonsense}. As a result,  \citet{huang-etal-2019-cosmos, talmor-etal-2019-commonsenseqa, sap2019atomic, west_symbolic_2021} introduce datasets to evaluate the extent to which LMs can reason over the knowledge they learned during pretraining.
In addition to implicit reasoning, \citet{talmor_leap--thought_2020} show a way in which LMs can systematically reason over both explicit input statements given to the model by a user and the implicit knowledge stored in the parameters of the model. However, despite succeeding in commonsense leaderboards, \citet{Merrill2021ProvableLO} suggest that those models fail to understand the underlying semantics leading them to commit trivial mistakes.


\section{Explainability \& Interpretability}
While the field of NLP has traditionally been guarded by the use of inherently explainable and interpretable techniques, the move from predominantly statistical NLP methods to black box neural models have only motivated the necessity to return to the study of explainability and interpretability.

While literature in NLP and computer science combine explainability and interpretability as similar, if not, identical aspects \cite{dosilovic_explainable_2018}, we aim to clarify these confusions by creating clear definitions for both terms. In short, we define \textbf{Interpretability} as the inspection of the inner workings of the model and understanding the reasoning behind model predictions by evaluating internal mechanisms. On the other hand, \textbf{Explainability} is a focus on the outward appearance of a model, namely whether the model's outputs are explainable in a post-hoc setting.

Since our main goal is to apply these \lmkb in practice, LMs that are not explainable and interpretable are often undeployable. For example, prior work has identified that BERT \cite{devlin-etal-2019-bert}, while powerful, is inherently opaque with respect to its inner workings \cite{rogers_primer_2020}. KBs, on the other hand, are far easier to read, as their fixed schemas are simpler to interpret. Having the same level of explainability inherent in LMs as we do in KBs would vastly improve their practical use. However, while KBs are naturally explainable, they do not inherently have interpretable qualities. So we make the distinction that, while we talk about both aspects with respect to LMs, we in tangent explain interpretability to cover all aspects of explainable AI in LMs.

\subsection{Interpretability}
Transformer models \cite{vaswani17attention, devlin-etal-2019-bert, Liu2019RoBERTa, Raffel2020ExploringTL} are composed of different building blocks: the encoder, decoder, self-attention, and more. However, we know little of how these building blocks work. Even as newer LMs are devised, some model capabilities remain uninterpreted, and problematic behaviors become evident even after a model has been in use for a long period of time. As a result, a deep understanding of the mechanisms that drive these models is imperative for long-term use and applied success in the real world.

\paragraph{Probing} Because little is known about how neural models function, many researchers opt to understand these models via probing. In its most basic form, one type of probing is done by way of the use of a simple linear classifier to associate internal representations with external properties \cite{belinkov-etal-2020-interpretability}. Probing allows researchers to answer questions about how the models function, their structure, or the decisions that these models make, especially regarding how these models learn linguistic structures \cite{tenney2018what,hewitt_designing_2019,hewitt-manning-2019-structural, belinkov-etal-2020-interpretability}.

\paragraph{Attention} With the introduction of attention-based LMs \cite{vaswani17attention}, researchers have attempted to use attention \cite{Bahdanau2015NeuralMT} to interpret the inner workings of the models. However, attention heads have been found to be fairly uninterpretable, as \citet{michel2019sixteen} has found that multiple attention heads have little impact on performance. Other transformer-based attention research has evaluated whether these models actually learn linguistic and syntactic structure and the relationship between them \cite{tenney2019bert, jawahar-etal-2019-bert, clark2019does, vig2019analyzing}.



However, many previous works have also contended whether attention can convey proper explanations \cite{jain2019attention, serrano2019attention, wiegreffe2019attention}. \citet{jain2019attention} and \citet{serrano2019attention} argue that attention cannot be used as a \textit{faithful} explanation for the model. On the other hand, \citet{wiegreffe2019attention} contradicts both statements by noting that there is a \textit{plausible} chance that attention could be correlated with the model. However, this claim is dependent on one's definition of explanation. \citet{mohankumar2020transparent} follows up prior work to note that the distribution of attention fails to fall on important words and strays to unimportant tokens. As a result, the definition of whether attention can adequately provide an explanation for the inner workings of LMs remains opaque. 

\paragraph{Transformer Mechanisms} We evaluate transformers and whether they are explainable via other mechanisms, such as the feedfoward layers \cite{zhao2021nonlinearity, geva2021transformer, meng2022locating}. \citet{zhao2021nonlinearity} propose a tool to measure non-linearities in LMs by taking into account geometry space of embeddings, finding that the non-linearity of the self-attention feedforward layers and MLPs of a LM follow similar patterns, but their functions are less known. \citet{geva2021transformer} extrapolate from this learned fact and find that feedforward layers in LMs are just key value memories; as a result \citet{meng2022locating} are able to use their method of causal tracing to locate the knowledge and use the key value pairs to access knowledge within the feedforward layers and make modifications to it.

\paragraph{Mechanistic Interpretability} Akin to reverse engineering software, if we could reverse engineer transformers, we could garner more understanding about the inner workings of these models. \citet{elhage2021mathematical} introduces mathematical conception as a way to understand the internal operations within. In addition, they also discover that attention heads can explain in-context learning within smaller models. Their promising results highlight the potential for further development of mathematical tools to understand computational patterns.

\paragraph{Causal tracing} Introduced in \citet{meng2022locating}, causal tracing is another form of accessing in which the model is traced for the path of information within. To do so, the model is run multiple times while corruptions are introduced to the system, then the system is restored to tease out which changes restored the original results. Their results show that it is possible to identify activations that are related to a model's factual predictions.

\subsection{Explainability}
The more we better explain the output of existing LMs, the better we can tailor these systems to real-world use. These explanations for model behavior could in turn be used to correct model shortcomings and improve help the end user gain trust in a system.

\paragraph{Influence Functions} One way to explain a black box model is through a technique known as an influence function. First introduced by \citet{hampel1974}, \citet{koh2020understanding} applies influence functions to neural models by using second-order optimization techniques; presenting a method that enable influence functions to be used to interpret model outputs.

\citet{han2020explaining} applies influence functions to LMs, finding that influence functions may be better suited for more complex tasks, such as natural language inference, despite these methods being computationally prohibitive \cite{pezeshkpour2021empirical}. Gradient-based and non-Hessian information \textit{influence attribution} methods \cite{pezeshkpour2021empirical} have been introduced to speed up computation. \cite{pezeshkpour2021combining} also introduce the combination of influence atrtibution with saliency maps to find artifacts in training data more accurately than using influence functions alone.

Influence functions have also been used for interactive debugging by way of using humans as feedback after interpreting their output \cite{zylberajch-etal-2021-hildif}. In this way, we show promise that there is a potential to integrate these explainability tools with humans in a more practical setting.

\paragraph{Explanations}
Researchers have introduced the potential for LMs to generate coherent explanations of their decisions. LMs such as T5 \cite{Raffel2020ExploringTL} are able to generate explanations that achieve state-of-the-art performance on explainability benchmarks \cite{Narang2020WT5TT}. Despite the push to use explanations as a way to improve explainability within LMs, these explanations are still inconsistent and fickle. \citet{Camburu2020MakeUY} show that by using an adversarial framework to interject modified inputs, they are able to show that LMs generate a large number of inconsistencies in their explanations.

Other methods have shown that LMs can learn to generate the reasoning process behind their decisions using prototype networks \cite{Schramowski21}, through highlighting fragments of the input text to justify the output \cite{Lei2016RationalizingNP}, or using human-provided explanations in the training process \cite{Camburu2018eSNLINL, paranjape-etal-2021-prompting}.
\section{Future Work \& Limitations}

In this paper, we review the literature with respect to five aspects that LMs need to be proficient at to qualify as KBs. However, despite these recent breakthroughs, the community still has a long way to go to enable the real-world deployment of LMs. For instance, pretrained LMs need explicit tuning on a consistency corpus \cite{elazar_measuring_2021} in order to behave similarly under different paraphrases. They are also sensitive to word-order, negation, priming, and patterns \cite{kassner2020negated} and are unreliable out-of-the-box. Furthermore, previous work demonstrates that there are theoretical limitations to transformers that prevents them from performing certain types of reasoning tasks \cite{hahn-2020-theoretical, Bhattamishra2020OnTP, helwe2021reasoning}. They also lack social intelligence \cite{Liang2021TowardsUA}, an understanding of time \cite{dhingra_time-aware_2021, lazaridou_mind_2021}, causality \cite{Li2021CausalBERTIC}, dealing with uncommon facts \cite{poerner-etal-2020-e, jiang-etal-2020-x} and counterfactual reasoning \cite{Feng2021EmpoweringLU}. We hope that by uncovering the limitations of LMs from the perspective of KBs, we can continue to motivate exploring intrinsically the positives of KBs and apply the knowledge to improving LMs.

\appendix

\section*{Acknowledgements}

Special thanks to Siddharth Verma for many helpful discussions and comments on the paper. Ahmed El-Kholy for the graphic in Figure \ref{fig:lmkb}.

\bibliography{anthology,custom}

\begin{thebibliography}{174}
\expandafter\ifx\csname natexlab\endcsname\relax\def\natexlab#1{#1}\fi

\bibitem[{Adolphs et~al.(2021)Adolphs, Dhuliawala, and
  Hofmann}]{Adolphs2021HowTQ}
Leonard Adolphs, Shehzaad Dhuliawala, and Thomas Hofmann. 2021.
\newblock How to query language models?
\newblock \emph{ArXiv}, abs/2108.01928.

\bibitem[{Aghajanyan et~al.(2022)Aghajanyan, Huang, Ross, Karpukhin, Xu, Goyal,
  Okhonko, Joshi, Ghosh, Lewis, and Zettlemoyer}]{Aghajanyan2022CM3AC}
Armen Aghajanyan, Bernie Huang, Candace Ross, Vladimir Karpukhin, Hu~Xu, Naman
  Goyal, Dmytro Okhonko, Mandar Joshi, Gargi Ghosh, Mike Lewis, and Luke
  Zettlemoyer. 2022.
\newblock Cm3: A causal masked multimodal model of the internet.
\newblock \emph{ArXiv}, abs/2201.07520.

\bibitem[{Aghajanyan et~al.(2021)Aghajanyan, Okhonko, Lewis, Joshi, Xu, Ghosh,
  and Zettlemoyer}]{aghajanyan_htlm_2021}
Armen Aghajanyan, Dmytro Okhonko, Mike Lewis, Mandar Joshi, Hu~Xu, Gargi Ghosh,
  and Luke Zettlemoyer. 2021.
\newblock \href {http://arxiv.org/abs/2107.06955} {{HTLM}: {Hyper}-{Text}
  {Pre}-{Training} and {Prompting} of {Language} {Models}}.
\newblock \emph{arXiv:2107.06955 [cs]}.
\newblock ArXiv: 2107.06955.

\bibitem[{Ahn et~al.(2017)Ahn, Choi, Pärnamaa, and Bengio}]{ahn2017neural}
Sungjin Ahn, Heeyoul Choi, Tanel Pärnamaa, and Yoshua Bengio. 2017.
\newblock \href {http://arxiv.org/abs/1608.00318} {A neural knowledge language
  model}.

\bibitem[{Andersen and Pretolani(2001)}]{andersen01}
KimAllan Andersen and Daniele Pretolani. 2001.
\newblock \href {https://doi.org/10.1023/A:1012332915908} {Easy cases of
  probabilistic satisfiability}.
\newblock \emph{Annals of Mathematics and Artificial Intelligence}, 33:69--91.

\bibitem[{Bahdanau et~al.(2015)Bahdanau, Cho, and
  Bengio}]{Bahdanau2015NeuralMT}
Dzmitry Bahdanau, Kyunghyun Cho, and Yoshua Bengio. 2015.
\newblock Neural machine translation by jointly learning to align and
  translate.
\newblock \emph{CoRR}, abs/1409.0473.

\bibitem[{Belinkov et~al.(2020)Belinkov, Gehrmann, and
  Pavlick}]{belinkov-etal-2020-interpretability}
Yonatan Belinkov, Sebastian Gehrmann, and Ellie Pavlick. 2020.
\newblock \href {https://doi.org/10.18653/v1/2020.acl-tutorials.1}
  {Interpretability and analysis in neural {NLP}}.
\newblock In \emph{Proceedings of the 58th Annual Meeting of the Association
  for Computational Linguistics: Tutorial Abstracts}, pages 1--5, Online.
  Association for Computational Linguistics.

\bibitem[{Bender et~al.(2021)Bender, Gebru, McMillan-Major, and
  Shmitchell}]{bender21}
Emily~M. Bender, Timnit Gebru, Angelina McMillan-Major, and Shmargaret
  Shmitchell. 2021.
\newblock \href {https://doi.org/10.1145/3442188.3445922} {On the dangers of
  stochastic parrots: Can language models be too big?}
\newblock In \emph{Proceedings of the 2021 ACM Conference on Fairness,
  Accountability, and Transparency}, FAccT '21, page 610–623, New York, NY,
  USA. Association for Computing Machinery.

\bibitem[{Bhagat and Hovy(2013)}]{bhagat13}
Rahul Bhagat and Eduard Hovy. 2013.
\newblock \href {https://doi.org/10.1162/COLI_a_00166} {{What Is a
  Paraphrase?}}
\newblock \emph{Computational Linguistics}, 39(3):463--472.

\bibitem[{Bhagavatula et~al.(2020)Bhagavatula, Bras, Malaviya, Sakaguchi,
  Holtzman, Rashkin, Downey, Yih, and Choi}]{Bhagavatula2020AbductiveCR}
Chandra Bhagavatula, Ronan~Le Bras, Chaitanya Malaviya, Keisuke Sakaguchi, Ari
  Holtzman, Hannah Rashkin, Doug Downey, Scott Yih, and Yejin Choi. 2020.
\newblock Abductive commonsense reasoning.
\newblock \emph{ArXiv}, abs/1908.05739.

\bibitem[{Bhattamishra et~al.(2020)Bhattamishra, Ahuja, and
  Goyal}]{Bhattamishra2020OnTP}
S.~Bhattamishra, Kabir Ahuja, and Navin Goyal. 2020.
\newblock On the practical ability of recurrent neural networks to recognize
  hierarchical languages.
\newblock \emph{ArXiv}, abs/2011.03965.

\bibitem[{Bowman et~al.(2015)Bowman, Angeli, Potts, and
  Manning}]{bowman-etal-2015-large}
Samuel~R. Bowman, Gabor Angeli, Christopher Potts, and Christopher~D. Manning.
  2015.
\newblock \href {https://doi.org/10.18653/v1/D15-1075} {A large annotated
  corpus for learning natural language inference}.
\newblock In \emph{Proceedings of the 2015 Conference on Empirical Methods in
  Natural Language Processing}, pages 632--642, Lisbon, Portugal. Association
  for Computational Linguistics.

\bibitem[{Brown et~al.(2020{\natexlab{a}})Brown, Mann, Ryder, Subbiah, Kaplan,
  Dhariwal, Neelakantan, Shyam, Sastry, Askell, Agarwal, Herbert-Voss, Krueger,
  Henighan, Child, Ramesh, Ziegler, Wu, Winter, Hesse, Chen, Sigler, Litwin,
  Gray, Chess, Clark, Berner, McCandlish, Radford, Sutskever, and
  Amodei}]{gpt3}
Tom Brown, Benjamin Mann, Nick Ryder, Melanie Subbiah, Jared~D Kaplan, Prafulla
  Dhariwal, Arvind Neelakantan, Pranav Shyam, Girish Sastry, Amanda Askell,
  Sandhini Agarwal, Ariel Herbert-Voss, Gretchen Krueger, Tom Henighan, Rewon
  Child, Aditya Ramesh, Daniel Ziegler, Jeffrey Wu, Clemens Winter, Chris
  Hesse, Mark Chen, Eric Sigler, Mateusz Litwin, Scott Gray, Benjamin Chess,
  Jack Clark, Christopher Berner, Sam McCandlish, Alec Radford, Ilya Sutskever,
  and Dario Amodei. 2020{\natexlab{a}}.
\newblock \href
  {https://proceedings.neurips.cc/paper/2020/file/1457c0d6bfcb4967418bfb8ac142f64a-Paper.pdf}
  {Language models are few-shot learners}.
\newblock In \emph{Advances in Neural Information Processing Systems},
  volume~33, pages 1877--1901. Curran Associates, Inc.

\bibitem[{Brown et~al.(2020{\natexlab{b}})Brown, Mann, Ryder, Subbiah, Kaplan,
  Dhariwal, Neelakantan, Shyam, Sastry, Askell, Agarwal, Herbert-Voss, Krueger,
  Henighan, Child, Ramesh, Ziegler, Wu, Winter, Hesse, Chen, Sigler, Litwin,
  Gray, Chess, Clark, Berner, McCandlish, Radford, Sutskever, and
  Amodei}]{brown2020language}
Tom~B. Brown, Benjamin Mann, Nick Ryder, Melanie Subbiah, Jared Kaplan,
  Prafulla Dhariwal, Arvind Neelakantan, Pranav Shyam, Girish Sastry, Amanda
  Askell, Sandhini Agarwal, Ariel Herbert-Voss, Gretchen Krueger, Tom Henighan,
  Rewon Child, Aditya Ramesh, Daniel~M. Ziegler, Jeffrey Wu, Clemens Winter,
  Christopher Hesse, Mark Chen, Eric Sigler, Mateusz Litwin, Scott Gray,
  Benjamin Chess, Jack Clark, Christopher Berner, Sam McCandlish, Alec Radford,
  Ilya Sutskever, and Dario Amodei. 2020{\natexlab{b}}.
\newblock \href {http://arxiv.org/abs/2005.14165} {Language models are few-shot
  learners}.

\bibitem[{Camburu et~al.(2018)Camburu, Rockt{\"a}schel, Lukasiewicz, and
  Blunsom}]{Camburu2018eSNLINL}
Oana-Maria Camburu, Tim Rockt{\"a}schel, Thomas Lukasiewicz, and Phil Blunsom.
  2018.
\newblock e-snli: Natural language inference with natural language
  explanations.
\newblock In \emph{NeurIPS}.

\bibitem[{Camburu et~al.(2020)Camburu, Shillingford, Minervini, Lukasiewicz,
  and Blunsom}]{Camburu2020MakeUY}
Oana-Maria Camburu, Brendan Shillingford, Pasquale Minervini, Thomas
  Lukasiewicz, and Phil Blunsom. 2020.
\newblock Make up your mind! adversarial generation of inconsistent natural
  language explanations.
\newblock In \emph{ACL}.

\bibitem[{Carlini et~al.(2021)Carlini, Tramer, Wallace, Jagielski,
  Herbert-Voss, Lee, Roberts, Brown, Song, Erlingsson, Oprea, and
  Raffel}]{carlini_extracting_2021}
Nicholas Carlini, Florian Tramer, Eric Wallace, Matthew Jagielski, Ariel
  Herbert-Voss, Katherine Lee, Adam Roberts, Tom Brown, Dawn Song, Ulfar
  Erlingsson, Alina Oprea, and Colin Raffel. 2021.
\newblock \href {http://arxiv.org/abs/2012.07805} {Extracting {Training} {Data}
  from {Large} {Language} {Models}}.
\newblock \emph{arXiv:2012.07805 [cs]}.
\newblock ArXiv: 2012.07805.

\bibitem[{Chen et~al.(2017)Chen, Liu, Yin, and Tang}]{chen-dialogue-2017}
Hongshen Chen, Xiaorui Liu, Dawei Yin, and Jiliang Tang. 2017.
\newblock \href {https://doi.org/10.1145/3166054.3166058} {A survey on dialogue
  systems}.
\newblock \emph{ACM SIGKDD Explorations Newsletter}, 19(2):25–35.

\bibitem[{Chung et~al.(2014)Chung, Çaglar G{\"u}lçehre, Cho, and
  Bengio}]{Chung2014EmpiricalEO}
Junyoung Chung, Çaglar G{\"u}lçehre, Kyunghyun Cho, and Yoshua Bengio. 2014.
\newblock Empirical evaluation of gated recurrent neural networks on sequence
  modeling.
\newblock \emph{ArXiv}, abs/1412.3555.

\bibitem[{Clark et~al.(2019)Clark, Khandelwal, Levy, and
  Manning}]{clark2019does}
Kevin Clark, Urvashi Khandelwal, Omer Levy, and Christopher~D. Manning. 2019.
\newblock \href {http://arxiv.org/abs/1906.04341} {What does bert look at? an
  analysis of bert's attention}.

\bibitem[{Clark et~al.(2020)Clark, Tafjord, and
  Richardson}]{Clark2020TransformersAS}
Peter Clark, Oyvind Tafjord, and Kyle Richardson. 2020.
\newblock Transformers as soft reasoners over language.
\newblock \emph{ArXiv}, abs/2002.05867.

\bibitem[{Colon-Hernandez et~al.(2021)Colon-Hernandez, Havasi, Alonso, Huggins,
  and Breazeal}]{colonhernandez2021combining}
Pedro Colon-Hernandez, Catherine Havasi, Jason Alonso, Matthew Huggins, and
  Cynthia Breazeal. 2021.
\newblock \href {http://arxiv.org/abs/2101.12294} {Combining pre-trained
  language models and structured knowledge}.

\bibitem[{Cui et~al.(2021)Cui, Cheng, Wu, and
  Zhang}]{cui-etal-2021-commonsense}
Leyang Cui, Sijie Cheng, Yu~Wu, and Yue Zhang. 2021.
\newblock \href {https://doi.org/10.18653/v1/2021.findings-acl.61} {On
  commonsense cues in {BERT} for solving commonsense tasks}.
\newblock In \emph{Findings of the Association for Computational Linguistics:
  ACL-IJCNLP 2021}, pages 683--693, Online. Association for Computational
  Linguistics.

\bibitem[{Da et~al.(2021)Da, Bras, Lu, Choi, and Bosselut}]{Da2021AnalyzingCE}
Jeff Da, Ronan~Le Bras, Ximing Lu, Yejin Choi, and Antoine Bosselut. 2021.
\newblock Analyzing commonsense emergence in few-shot knowledge models.

\bibitem[{Danilevsky et~al.(2020)Danilevsky, Qian, Aharonov, Katsis, Kawas, and
  Sen}]{danilevsky_survey_2020}
Marina Danilevsky, Kun Qian, Ranit Aharonov, Yannis Katsis, Ban Kawas, and
  Prithviraj Sen. 2020.
\newblock \href {http://arxiv.org/abs/2010.00711} {A {Survey} of the {State} of
  {Explainable} {AI} for {Natural} {Language} {Processing}}.
\newblock \emph{arXiv:2010.00711 [cs]}.
\newblock ArXiv: 2010.00711.

\bibitem[{Davison et~al.(2019)Davison, Feldman, and
  Rush}]{davison-etal-2019-commonsense}
Joe Davison, Joshua Feldman, and Alexander Rush. 2019.
\newblock \href {https://doi.org/10.18653/v1/D19-1109} {Commonsense knowledge
  mining from pretrained models}.
\newblock In \emph{Proceedings of the 2019 Conference on Empirical Methods in
  Natural Language Processing and the 9th International Joint Conference on
  Natural Language Processing (EMNLP-IJCNLP)}, pages 1173--1178, Hong Kong,
  China. Association for Computational Linguistics.

\bibitem[{De~Cao et~al.(2021)De~Cao, Aziz, and Titov}]{de_cao_editing_2021}
Nicola De~Cao, Wilker Aziz, and Ivan Titov. 2021.
\newblock \href {http://arxiv.org/abs/2104.08164} {Editing {Factual}
  {Knowledge} in {Language} {Models}}.
\newblock \emph{arXiv:2104.08164 [cs]}.
\newblock ArXiv: 2104.08164.

\bibitem[{de~Jong et~al.(2021)de~Jong, Zemlyanskiy, FitzGerald, Sha, and
  Cohen}]{dejong2021mention}
Michiel de~Jong, Yury Zemlyanskiy, Nicholas FitzGerald, Fei Sha, and William
  Cohen. 2021.
\newblock \href {http://arxiv.org/abs/2110.06176} {Mention memory:
  incorporating textual knowledge into transformers through entity mention
  attention}.

\bibitem[{de~Masson~d'Autume et~al.(2019)de~Masson~d'Autume, Ruder, Kong, and
  Yogatama}]{dAutume2019EpisodicMI}
Cyprien de~Masson~d'Autume, Sebastian Ruder, Lingpeng Kong, and Dani Yogatama.
  2019.
\newblock Episodic memory in lifelong language learning.
\newblock In \emph{NeurIPS}.

\bibitem[{Devlin et~al.(2019)Devlin, Chang, Lee, and
  Toutanova}]{devlin-etal-2019-bert}
Jacob Devlin, Ming-Wei Chang, Kenton Lee, and Kristina Toutanova. 2019.
\newblock \href {https://doi.org/10.18653/v1/N19-1423} {{BERT}: Pre-training of
  deep bidirectional transformers for language understanding}.
\newblock In \emph{Proceedings of the 2019 Conference of the North {A}merican
  Chapter of the Association for Computational Linguistics: Human Language
  Technologies, Volume 1 (Long and Short Papers)}, pages 4171--4186,
  Minneapolis, Minnesota. Association for Computational Linguistics.

\bibitem[{Dhingra et~al.(2021)Dhingra, Cole, Eisenschlos, Gillick, Eisenstein,
  and Cohen}]{dhingra_time-aware_2021}
Bhuwan Dhingra, Jeremy~R. Cole, Julian~Martin Eisenschlos, Daniel Gillick,
  Jacob Eisenstein, and William~W. Cohen. 2021.
\newblock \href {http://arxiv.org/abs/2106.15110} {Time-{Aware} {Language}
  {Models} as {Temporal} {Knowledge} {Bases}}.
\newblock \emph{arXiv:2106.15110 [cs]}.
\newblock ArXiv: 2106.15110.

\bibitem[{Dhingra et~al.(2020)Dhingra, Zaheer, Balachandran, Neubig,
  Salakhutdinov, and Cohen}]{dhingra2020differentiable}
Bhuwan Dhingra, Manzil Zaheer, Vidhisha Balachandran, Graham Neubig, Ruslan
  Salakhutdinov, and William~W. Cohen. 2020.
\newblock \href {http://arxiv.org/abs/2002.10640} {Differentiable reasoning
  over a virtual knowledge base}.

\bibitem[{Dong et~al.(2019)Dong, Yang, Wang, Wei, Liu, Wang, Gao, Zhou, and
  Hon}]{dong_NEURIPS2019_c20bb2d9}
Li~Dong, Nan Yang, Wenhui Wang, Furu Wei, Xiaodong Liu, Russell Wang, Jianfeng
  Gao, Ming Zhou, and Hsiao-Wuen Hon. 2019.
\newblock \href
  {https://proceedings.neurips.cc/paper/2019/file/c20bb2d9a50d5ac1f713f8b34d9aac5a-Paper.pdf}
  {Unified language model pre-training for natural language understanding and
  generation}.
\newblock In \emph{Advances in Neural Information Processing Systems},
  volume~32. Curran Associates, Inc.

\bibitem[{Došilović et~al.(2018)Došilović, Brčić, and
  Hlupić}]{dosilovic_explainable_2018}
Filip~Karlo Došilović, Mario Brčić, and Nikica Hlupić. 2018.
\newblock \href {https://doi.org/10.23919/MIPRO.2018.8400040} {Explainable
  artificial intelligence: A survey}.
\newblock In \emph{2018 41st International Convention on Information and
  Communication Technology, Electronics and Microelectronics (MIPRO)}, pages
  0210--0215.

\bibitem[{Elazar et~al.(2021)Elazar, Kassner, Ravfogel, Ravichander, Hovy,
  Schütze, and Goldberg}]{elazar_measuring_2021}
Yanai Elazar, Nora Kassner, Shauli Ravfogel, Abhilasha Ravichander, Eduard
  Hovy, Hinrich Schütze, and Yoav Goldberg. 2021.
\newblock \href {http://arxiv.org/abs/2102.01017} {Measuring and {Improving}
  {Consistency} in {Pretrained} {Language} {Models}}.
\newblock \emph{arXiv:2102.01017 [cs]}.
\newblock ArXiv: 2102.01017.

\bibitem[{Elhage et~al.(2021)Elhage, Nanda, Olsson, Henighan, Joseph, Mann,
  Askell, Bai, Chen, Conerly, DasSarma, Drain, Ganguli, Hatfield-Dodds,
  Hernandez, Jones, Kernion, Lovitt, Ndousse, Amodei, Brown, Clark, Kaplan,
  McCandlish, and Olah}]{elhage2021mathematical}
Nelson Elhage, Neel Nanda, Catherine Olsson, Tom Henighan, Nicholas Joseph, Ben
  Mann, Amanda Askell, Yuntao Bai, Anna Chen, Tom Conerly, Nova DasSarma, Dawn
  Drain, Deep Ganguli, Zac Hatfield-Dodds, Danny Hernandez, Andy Jones, Jackson
  Kernion, Liane Lovitt, Kamal Ndousse, Dario Amodei, Tom Brown, Jack Clark,
  Jared Kaplan, Sam McCandlish, and Chris Olah. 2021.
\newblock A mathematical framework for transformer circuits.
\newblock \emph{Transformer Circuits Thread}.
\newblock Https://transformer-circuits.pub/2021/framework/index.html.

\bibitem[{Ettinger(2020)}]{Ettinger20}
Allyson Ettinger. 2020.
\newblock \href {https://doi.org/10.1162/tacl_a_00298} {{What BERT Is Not:
  Lessons from a New Suite of Psycholinguistic Diagnostics for Language
  Models}}.
\newblock \emph{Transactions of the Association for Computational Linguistics},
  8:34--48.

\bibitem[{Feng et~al.(2021)Feng, Zhang, He, Zhang, and
  Chua}]{Feng2021EmpoweringLU}
Fuli Feng, Jizhi Zhang, Xiangnan He, Hanwang Zhang, and Tat-Seng Chua. 2021.
\newblock Empowering language understanding with counterfactual reasoning.
\newblock \emph{ArXiv}, abs/2106.03046.

\bibitem[{F{\'{e}}vry et~al.(2020)F{\'{e}}vry, Soares, FitzGerald, Choi, and
  Kwiatkowski}]{entities-as-experts}
Thibault F{\'{e}}vry, Livio~Baldini Soares, Nicholas FitzGerald, Eunsol Choi,
  and Tom Kwiatkowski. 2020.
\newblock \href {http://arxiv.org/abs/2004.07202} {Entities as experts: Sparse
  memory access with entity supervision}.
\newblock \emph{CoRR}, abs/2004.07202.

\bibitem[{F{\'e}vry et~al.(2020)F{\'e}vry, Soares, FitzGerald, Choi, and
  Kwiatkowski}]{Fvry2020EntitiesAE}
Thibault F{\'e}vry, Livio~Baldini Soares, Nicholas FitzGerald, Eunsol Choi, and
  Tom Kwiatkowski. 2020.
\newblock Entities as experts: Sparse memory access with entity supervision.
\newblock In \emph{EMNLP}.

\bibitem[{Finn et~al.(2017)Finn, Abbeel, and Levine}]{Finn2017ModelAgnosticMF}
Chelsea Finn, P.~Abbeel, and Sergey Levine. 2017.
\newblock Model-agnostic meta-learning for fast adaptation of deep networks.
\newblock In \emph{ICML}.

\bibitem[{Gao et~al.(2021)Gao, Fisch, and Chen}]{gao2021making}
Tianyu Gao, Adam Fisch, and Danqi Chen. 2021.
\newblock \href {http://arxiv.org/abs/2012.15723} {Making pre-trained language
  models better few-shot learners}.

\bibitem[{Gehman et~al.(2020)Gehman, Gururangan, Sap, Choi, and
  Smith}]{Gehman2020RealToxicityPromptsEN}
Samuel Gehman, Suchin Gururangan, Maarten Sap, Yejin Choi, and Noah~A. Smith.
  2020.
\newblock Realtoxicityprompts: Evaluating neural toxic degeneration in language
  models.
\newblock \emph{ArXiv}, abs/2009.11462.

\bibitem[{Geva et~al.(2021)Geva, Schuster, Berant, and
  Levy}]{geva2021transformer}
Mor Geva, Roei Schuster, Jonathan Berant, and Omer Levy. 2021.
\newblock \href {http://arxiv.org/abs/2012.14913} {Transformer feed-forward
  layers are key-value memories}.

\bibitem[{Goldberg(2019)}]{Goldberg2019AssessingBS}
Yoav Goldberg. 2019.
\newblock Assessing bert's syntactic abilities.
\newblock \emph{ArXiv}, abs/1901.05287.

\bibitem[{Gontier et~al.(2020)Gontier, Sinha, Reddy, and
  Pal}]{Gontier2020MeasuringSG}
Nicolas Gontier, Koustuv Sinha, Siva Reddy, and Christopher~Joseph Pal. 2020.
\newblock Measuring systematic generalization in neural proof generation with
  transformers.
\newblock \emph{ArXiv}, abs/2009.14786.

\bibitem[{Graves et~al.(2014)Graves, Wayne, and Danihelka}]{graves2014neural}
Alex Graves, Greg Wayne, and Ivo Danihelka. 2014.
\newblock \href {http://arxiv.org/abs/1410.5401} {Neural turing machines}.

\bibitem[{Guu et~al.(2020)Guu, Lee, Tung, Pasupat, and Chang}]{guu_realm_2020}
Kelvin Guu, Kenton Lee, Zora Tung, Panupong Pasupat, and Ming-Wei Chang. 2020.
\newblock \href {http://arxiv.org/abs/2002.08909} {{REALM}:
  {Retrieval}-{Augmented} {Language} {Model} {Pre}-{Training}}.
\newblock \emph{arXiv:2002.08909 [cs]}.
\newblock ArXiv: 2002.08909.

\bibitem[{Ha et~al.(2017)Ha, Dai, and Le}]{HyperNetworks}
David Ha, Andrew Dai, and Quoc~V. Le. 2017.
\newblock \href {https://openreview.net/pdf?id=rkpACe1lx} {Hypernetworks}.

\bibitem[{Hahn(2020)}]{hahn-2020-theoretical}
Michael Hahn. 2020.
\newblock \href {https://doi.org/10.1162/tacl_a_00306} {Theoretical limitations
  of self-attention in neural sequence models}.
\newblock \emph{Transactions of the Association for Computational Linguistics},
  8:156--171.

\bibitem[{Hampel(1974)}]{hampel1974}
Frank~R. Hampel. 1974.
\newblock \href {https://doi.org/10.1080/01621459.1974.10482962} {The influence
  curve and its role in robust estimation}.
\newblock \emph{Journal of the American Statistical Association},
  69(346):383--393.

\bibitem[{Han et~al.(2020)Han, Wallace, and Tsvetkov}]{han2020explaining}
Xiaochuang Han, Byron~C. Wallace, and Yulia Tsvetkov. 2020.
\newblock \href {http://arxiv.org/abs/2005.06676} {Explaining black box
  predictions and unveiling data artifacts through influence functions}.

\bibitem[{Han et~al.(2021)Han, Zhao, Ding, Liu, and Sun}]{han2021ptr}
Xu~Han, Weilin Zhao, Ning Ding, Zhiyuan Liu, and Maosong Sun. 2021.
\newblock \href {http://arxiv.org/abs/2105.11259} {Ptr: Prompt tuning with
  rules for text classification}.

\bibitem[{Hase et~al.(2021)Hase, Diab, Celikyilmaz, Li, Kozareva, Stoyanov,
  Bansal, and Iyer}]{hase_language_2021}
Peter Hase, Mona Diab, Asli Celikyilmaz, Xian Li, Zornitsa Kozareva, Veselin
  Stoyanov, Mohit Bansal, and Srinivasan Iyer. 2021.
\newblock \href {http://arxiv.org/abs/2111.13654} {Do {Language} {Models}
  {Have} {Beliefs}? {Methods} for {Detecting}, {Updating}, and {Visualizing}
  {Model} {Beliefs}}.
\newblock \emph{arXiv:2111.13654 [cs]}.
\newblock ArXiv: 2111.13654.

\bibitem[{He et~al.(2020)He, Zhou, Xiao, Jiang, Liu, Yuan, and
  Xu}]{he-etal-2020-bert}
Bin He, Di~Zhou, Jinghui Xiao, Xin Jiang, Qun Liu, Nicholas~Jing Yuan, and Tong
  Xu. 2020.
\newblock \href {https://doi.org/10.18653/v1/2020.findings-emnlp.207}
  {{BERT}-{MK}: Integrating graph contextualized knowledge into pre-trained
  language models}.
\newblock In \emph{Findings of the Association for Computational Linguistics:
  EMNLP 2020}, pages 2281--2290, Online. Association for Computational
  Linguistics.

\bibitem[{Heinzerling and Inui(2021)}]{heinzerling_language_2021}
Benjamin Heinzerling and Kentaro Inui. 2021.
\newblock \href {http://arxiv.org/abs/2008.09036} {Language {Models} as
  {Knowledge} {Bases}: {On} {Entity} {Representations}, {Storage} {Capacity},
  and {Paraphrased} {Queries}}.
\newblock \emph{arXiv:2008.09036 [cs]}.
\newblock ArXiv: 2008.09036.

\bibitem[{Helwe et~al.(2021)Helwe, Clavel, and Suchanek}]{helwe2021reasoning}
Chadi Helwe, Chlo{\'e} Clavel, and Fabian~M. Suchanek. 2021.
\newblock \href {https://openreview.net/forum?id=Ozp1WrgtF5_} {Reasoning with
  transformer-based models: Deep learning, but shallow reasoning}.
\newblock In \emph{3rd Conference on Automated Knowledge Base Construction}.

\bibitem[{Hewitt and Liang(2019)}]{hewitt_designing_2019}
John Hewitt and Percy Liang. 2019.
\newblock \href {http://arxiv.org/abs/1909.03368} {Designing and {Interpreting}
  {Probes} with {Control} {Tasks}}.
\newblock \emph{arXiv:1909.03368 [cs]}.
\newblock ArXiv: 1909.03368.

\bibitem[{Hewitt and Manning(2019)}]{hewitt-manning-2019-structural}
John Hewitt and Christopher~D. Manning. 2019.
\newblock \href {https://doi.org/10.18653/v1/N19-1419} {{A} structural probe
  for finding syntax in word representations}.
\newblock In \emph{Proceedings of the 2019 Conference of the North {A}merican
  Chapter of the Association for Computational Linguistics: Human Language
  Technologies, Volume 1 (Long and Short Papers)}, pages 4129--4138,
  Minneapolis, Minnesota. Association for Computational Linguistics.

\bibitem[{Hochreiter and Schmidhuber(1997)}]{hochreiter1997long}
Sepp Hochreiter and J{\"u}rgen Schmidhuber. 1997.
\newblock Long short-term memory.
\newblock \emph{Neural computation}, 9(8):1735--1780.

\bibitem[{Houlsby et~al.(2019)Houlsby, Giurgiu, Jastrzebski, Morrone,
  de~Laroussilhe, Gesmundo, Attariyan, and
  Gelly}]{Houlsby2019ParameterEfficientTL}
Neil Houlsby, Andrei Giurgiu, Stanislaw Jastrzebski, Bruna Morrone, Quentin
  de~Laroussilhe, Andrea Gesmundo, Mona Attariyan, and Sylvain Gelly. 2019.
\newblock Parameter-efficient transfer learning for nlp.
\newblock In \emph{ICML}.

\bibitem[{Huang et~al.(2019)Huang, Le~Bras, Bhagavatula, and
  Choi}]{huang-etal-2019-cosmos}
Lifu Huang, Ronan Le~Bras, Chandra Bhagavatula, and Yejin Choi. 2019.
\newblock \href {https://doi.org/10.18653/v1/D19-1243} {Cosmos {QA}: Machine
  reading comprehension with contextual commonsense reasoning}.
\newblock In \emph{Proceedings of the 2019 Conference on Empirical Methods in
  Natural Language Processing and the 9th International Joint Conference on
  Natural Language Processing (EMNLP-IJCNLP)}, pages 2391--2401, Hong Kong,
  China. Association for Computational Linguistics.

\bibitem[{Huang et~al.(2022)Huang, Abbeel, Pathak, and
  Mordatch}]{Huang2022LanguageMA}
Wenlong Huang, P.~Abbeel, Deepak Pathak, and Igor Mordatch. 2022.
\newblock Language models as zero-shot planners: Extracting actionable
  knowledge for embodied agents.
\newblock \emph{ArXiv}, abs/2201.07207.

\bibitem[{Jain and Wallace(2019)}]{jain2019attention}
Sarthak Jain and Byron~C. Wallace. 2019.
\newblock \href {http://arxiv.org/abs/1902.10186} {Attention is not
  explanation}.

\bibitem[{Jawahar et~al.(2019)Jawahar, Sagot, and
  Seddah}]{jawahar-etal-2019-bert}
Ganesh Jawahar, Beno{\^\i}t Sagot, and Djam{\'e} Seddah. 2019.
\newblock \href {https://doi.org/10.18653/v1/P19-1356} {What does {BERT} learn
  about the structure of language?}
\newblock In \emph{Proceedings of the 57th Annual Meeting of the Association
  for Computational Linguistics}, pages 3651--3657, Florence, Italy.
  Association for Computational Linguistics.

\bibitem[{Ji et~al.(2020)Ji, Ke, Huang, Wei, Zhu, and Huang}]{ji2020language}
Haozhe Ji, Pei Ke, Shaohan Huang, Furu Wei, Xiaoyan Zhu, and Minlie Huang.
  2020.
\newblock \href {http://arxiv.org/abs/2009.11692} {Language generation with
  multi-hop reasoning on commonsense knowledge graph}.

\bibitem[{Ji et~al.(2021)Ji, Pan, Cambria, Marttinen, and Yu}]{ji_survey_2021}
Shaoxiong Ji, Shirui Pan, Erik Cambria, Pekka Marttinen, and Philip~S. Yu.
  2021.
\newblock \href {https://doi.org/10.1109/TNNLS.2021.3070843} {A {Survey} on
  {Knowledge} {Graphs}: {Representation}, {Acquisition} and {Applications}}.
\newblock \emph{arXiv:2002.00388 [cs]}.
\newblock ArXiv: 2002.00388.

\bibitem[{Jiang et~al.(2020{\natexlab{a}})Jiang, Anastasopoulos, Araki, Ding,
  and Neubig}]{jiang-etal-2020-x}
Zhengbao Jiang, Antonios Anastasopoulos, Jun Araki, Haibo Ding, and Graham
  Neubig. 2020{\natexlab{a}}.
\newblock \href {https://doi.org/10.18653/v1/2020.emnlp-main.479} {{X}-{FACTR}:
  Multilingual factual knowledge retrieval from pretrained language models}.
\newblock In \emph{Proceedings of the 2020 Conference on Empirical Methods in
  Natural Language Processing (EMNLP)}, pages 5943--5959, Online. Association
  for Computational Linguistics.

\bibitem[{Jiang et~al.(2020{\natexlab{b}})Jiang, Xu, Araki, and
  Neubig}]{jiang_how_2020}
Zhengbao Jiang, Frank~F. Xu, Jun Araki, and Graham Neubig. 2020{\natexlab{b}}.
\newblock \href {http://arxiv.org/abs/1911.12543} {How {Can} {We} {Know} {What}
  {Language} {Models} {Know}?}
\newblock \emph{arXiv:1911.12543 [cs]}.
\newblock ArXiv: 1911.12543.

\bibitem[{Kassner and Schütze(2020)}]{kassner2020negated}
Nora Kassner and Hinrich Schütze. 2020.
\newblock \href {http://arxiv.org/abs/1911.03343} {Negated and misprimed probes
  for pretrained language models: Birds can talk, but cannot fly}.

\bibitem[{Kassner et~al.(2021)Kassner, Tafjord, Schütze, and
  Clark}]{kassner_beliefbank_2021}
Nora Kassner, Oyvind Tafjord, Hinrich Schütze, and Peter Clark. 2021.
\newblock \href {http://arxiv.org/abs/2109.14723} {{BeliefBank}: {Adding}
  {Memory} to a {Pre}-{Trained} {Language} {Model} for a {Systematic} {Notion}
  of {Belief}}.
\newblock \emph{arXiv:2109.14723 [cs]}.
\newblock ArXiv: 2109.14723.

\bibitem[{Kim(2014)}]{kim-2014-convolutional}
Yoon Kim. 2014.
\newblock \href {https://doi.org/10.3115/v1/D14-1181} {Convolutional neural
  networks for sentence classification}.
\newblock In \emph{Proceedings of the 2014 Conference on Empirical Methods in
  Natural Language Processing ({EMNLP})}, pages 1746--1751, Doha, Qatar.
  Association for Computational Linguistics.

\bibitem[{Koh and Liang(2020)}]{koh2020understanding}
Pang~Wei Koh and Percy Liang. 2020.
\newblock \href {http://arxiv.org/abs/1703.04730} {Understanding black-box
  predictions via influence functions}.

\bibitem[{Lakkaraju et~al.(2022)Lakkaraju, Slack, Chen, Tan, and
  Singh}]{Lakkaraju2022RethinkingEA}
Himabindu Lakkaraju, Dylan Slack, Yuxin Chen, Chenhao Tan, and Sameer Singh.
  2022.
\newblock Rethinking explainability as a dialogue: A practitioner's
  perspective.
\newblock \emph{ArXiv}, abs/2202.01875.

\bibitem[{Lazaridou et~al.(2021)Lazaridou, Kuncoro, Gribovskaya, Agrawal,
  Liska, Terzi, Gimenez, d'Autume, Kocisky, Ruder, Yogatama, Cao, Young, and
  Blunsom}]{lazaridou_mind_2021}
Angeliki Lazaridou, Adhiguna Kuncoro, Elena Gribovskaya, Devang Agrawal, Adam
  Liska, Tayfun Terzi, Mai Gimenez, Cyprien de~Masson d'Autume, Tomas Kocisky,
  Sebastian Ruder, Dani Yogatama, Kris Cao, Susannah Young, and Phil Blunsom.
  2021.
\newblock \href {http://arxiv.org/abs/2102.01951} {Mind the {Gap}: {Assessing}
  {Temporal} {Generalization} in {Neural} {Language} {Models}}.
\newblock \emph{arXiv:2102.01951 [cs]}.
\newblock ArXiv: 2102.01951.

\bibitem[{Lei et~al.(2016)Lei, Barzilay, and Jaakkola}]{Lei2016RationalizingNP}
Tao Lei, Regina Barzilay, and T.~Jaakkola. 2016.
\newblock Rationalizing neural predictions.
\newblock In \emph{EMNLP}.

\bibitem[{Lester et~al.(2021)Lester, Al-Rfou, and Constant}]{lester2021power}
Brian Lester, Rami Al-Rfou, and Noah Constant. 2021.
\newblock \href {http://arxiv.org/abs/2104.08691} {The power of scale for
  parameter-efficient prompt tuning}.

\bibitem[{Levy et~al.(2017)Levy, Seo, Choi, and
  Zettlemoyer}]{levy-etal-2017-zero}
Omer Levy, Minjoon Seo, Eunsol Choi, and Luke Zettlemoyer. 2017.
\newblock \href {https://doi.org/10.18653/v1/K17-1034} {Zero-shot relation
  extraction via reading comprehension}.
\newblock In \emph{Proceedings of the 21st Conference on Computational Natural
  Language Learning ({C}o{NLL} 2017)}, pages 333--342, Vancouver, Canada.
  Association for Computational Linguistics.

\bibitem[{Lewis et~al.(2020)Lewis, Liu, Goyal, Ghazvininejad, Mohamed, Levy,
  Stoyanov, and Zettlemoyer}]{lewis-etal-2020-bart}
Mike Lewis, Yinhan Liu, Naman Goyal, Marjan Ghazvininejad, Abdelrahman Mohamed,
  Omer Levy, Veselin Stoyanov, and Luke Zettlemoyer. 2020.
\newblock \href {https://doi.org/10.18653/v1/2020.acl-main.703} {{BART}:
  Denoising sequence-to-sequence pre-training for natural language generation,
  translation, and comprehension}.
\newblock In \emph{Proceedings of the 58th Annual Meeting of the Association
  for Computational Linguistics}, pages 7871--7880, Online. Association for
  Computational Linguistics.

\bibitem[{Lewis et~al.(2021)Lewis, Perez, Piktus, Petroni, Karpukhin, Goyal,
  Küttler, Lewis, Yih, Rocktäschel, Riedel, and
  Kiela}]{lewis_retrieval-augmented_2021}
Patrick Lewis, Ethan Perez, Aleksandra Piktus, Fabio Petroni, Vladimir
  Karpukhin, Naman Goyal, Heinrich Küttler, Mike Lewis, Wen-tau Yih, Tim
  Rocktäschel, Sebastian Riedel, and Douwe Kiela. 2021.
\newblock \href {http://arxiv.org/abs/2005.11401} {Retrieval-{Augmented}
  {Generation} for {Knowledge}-{Intensive} {NLP} {Tasks}}.
\newblock \emph{arXiv:2005.11401 [cs]}.
\newblock ArXiv: 2005.11401.

\bibitem[{Li and Liang(2021)}]{li2021prefixtuning}
Xiang~Lisa Li and Percy Liang. 2021.
\newblock \href {http://arxiv.org/abs/2101.00190} {Prefix-tuning: Optimizing
  continuous prompts for generation}.

\bibitem[{Li et~al.(2021)Li, Ding, Liao, Liu, and Qin}]{Li2021CausalBERTIC}
Zhongyang Li, Xiao Ding, Kuo Liao, Ting Liu, and Bing Qin. 2021.
\newblock Causalbert: Injecting causal knowledge into pre-trained models with
  minimal supervision.
\newblock \emph{ArXiv}, abs/2107.09852.

\bibitem[{Liang et~al.(2021)Liang, Wu, Morency, and
  Salakhutdinov}]{Liang2021TowardsUA}
Paul~Pu Liang, Chiyu Wu, Louis-Philippe Morency, and Ruslan Salakhutdinov.
  2021.
\newblock Towards understanding and mitigating social biases in language
  models.
\newblock In \emph{ICML}.

\bibitem[{Lin et~al.(2021{\natexlab{a}})Lin, Hilton, and
  Evans}]{Lin2021TruthfulQAMH}
Stephanie~C. Lin, Jacob Hilton, and Owain Evans. 2021{\natexlab{a}}.
\newblock Truthfulqa: Measuring how models mimic human falsehoods.
\newblock \emph{ArXiv}, abs/2109.07958.

\bibitem[{Lin et~al.(2021{\natexlab{b}})Lin, Mihaylov, Artetxe, Wang, Chen,
  Simig, Ott, Goyal, Bhosale, Du, Pasunuru, Shleifer, Koura, Chaudhary, O'Horo,
  Wang, Zettlemoyer, Kozareva, Diab, Stoyanov, and Li}]{Lin2021FewshotLW}
Xi~Victoria Lin, Todor Mihaylov, Mikel Artetxe, Tianlu Wang, Shuohui Chen,
  Daniel Simig, Myle Ott, Naman Goyal, Shruti Bhosale, Jingfei Du, Ramakanth
  Pasunuru, Sam Shleifer, Punit~Singh Koura, Vishrav Chaudhary, Brian O'Horo,
  Jeff Wang, Luke Zettlemoyer, Zornitsa Kozareva, Mona Diab, Ves Stoyanov, and
  Xian Li. 2021{\natexlab{b}}.
\newblock Few-shot learning with multilingual language models.
\newblock \emph{ArXiv}, abs/2112.10668.

\bibitem[{Lipton(2018)}]{Lipton2018TheMO}
Zachary~Chase Lipton. 2018.
\newblock The mythos of model interpretability.
\newblock \emph{Queue}, 16:31 -- 57.

\bibitem[{Liu et~al.(2021{\natexlab{a}})Liu, Shen, Zhang, Dolan, Carin, and
  Chen}]{liu2021makes}
Jiachang Liu, Dinghan Shen, Yizhe Zhang, Bill Dolan, Lawrence Carin, and Weizhu
  Chen. 2021{\natexlab{a}}.
\newblock \href {http://arxiv.org/abs/2101.06804} {What makes good in-context
  examples for gpt-$3$?}

\bibitem[{Liu et~al.(2021{\natexlab{b}})Liu, Wang, Kasai, Hajishirzi, and
  Smith}]{liu2021probing}
Leo~Z. Liu, Yizhong Wang, Jungo Kasai, Hannaneh Hajishirzi, and Noah~A. Smith.
  2021{\natexlab{b}}.
\newblock \href {http://arxiv.org/abs/2104.07885} {Probing across time: What
  does roberta know and when?}

\bibitem[{Liu et~al.(2021{\natexlab{c}})Liu, Li, He, Bing, Joty, and
  Si}]{liu_knowledge_2021}
Linlin Liu, Xin Li, Ruidan He, Lidong Bing, Shafiq Joty, and Luo Si.
  2021{\natexlab{c}}.
\newblock \href {http://arxiv.org/abs/2111.10962} {Knowledge {Based}
  {Multilingual} {Language} {Model}}.
\newblock \emph{arXiv:2111.10962 [cs]}.
\newblock ArXiv: 2111.10962.

\bibitem[{Liu et~al.(2021{\natexlab{d}})Liu, Yuan, Fu, Jiang, Hayashi, and
  Neubig}]{Liu2021PretrainPA}
Pengfei Liu, Weizhe Yuan, Jinlan Fu, Zhengbao Jiang, Hiroaki Hayashi, and
  Graham Neubig. 2021{\natexlab{d}}.
\newblock Pre-train, prompt, and predict: A systematic survey of prompting
  methods in natural language processing.
\newblock \emph{ArXiv}, abs/2107.13586.

\bibitem[{Liu et~al.(2022)Liu, Yogatama, and Blunsom}]{Liu2022RelationalMA}
Qi~Liu, Dani Yogatama, and Phil Blunsom. 2022.
\newblock Relational memory augmented language models.
\newblock \emph{ArXiv}, abs/2201.09680.

\bibitem[{Liu et~al.(2020)Liu, Zhou, Zhao, Wang, Ju, Deng, and
  Wang}]{Liu2020KBERTEL}
Weijie Liu, Peng Zhou, Zhe Zhao, Zhiruo Wang, Qi~Ju, Haotang Deng, and Ping
  Wang. 2020.
\newblock K-bert: Enabling language representation with knowledge graph.
\newblock In \emph{AAAI}.

\bibitem[{Liu et~al.(2019)Liu, Ott, Goyal, Du, Joshi, Chen, Levy, Lewis,
  Zettlemoyer, and Stoyanov}]{Liu2019RoBERTa}
Yinhan Liu, Myle Ott, Naman Goyal, Jingfei Du, Mandar Joshi, Danqi Chen, Omer
  Levy, M.~Lewis, Luke Zettlemoyer, and Veselin Stoyanov. 2019.
\newblock Roberta: A robustly optimized bert pretraining approach.
\newblock \emph{ArXiv}, abs/1907.11692.

\bibitem[{Logan et~al.(2021)Logan, Balavzevi'c, Wallace, Petroni, Singh, and
  Riedel}]{LoganIV2021CuttingDO}
Robert~L Logan, Ivana Balavzevi'c, Eric Wallace, Fabio Petroni, Sameer Singh,
  and Sebastian Riedel. 2021.
\newblock Cutting down on prompts and parameters: Simple few-shot learning with
  language models.
\newblock \emph{ArXiv}, abs/2106.13353.

\bibitem[{Logan et~al.(2019)Logan, Liu, Peters, Gardner, and
  Singh}]{logan2019baracks}
Robert~L. Logan, Nelson~F. Liu, Matthew~E. Peters, Matt Gardner, and Sameer
  Singh. 2019.
\newblock \href {http://arxiv.org/abs/1906.07241} {Barack's wife hillary: Using
  knowledge-graphs for fact-aware language modeling}.

\bibitem[{Maillard et~al.(2021)Maillard, Karpukhin, Petroni, tau Yih, Oğuz,
  Stoyanov, and Ghosh}]{Maillard2021MultitaskRF}
Jean Maillard, Vladimir Karpukhin, Fabio Petroni, Wen tau Yih, Barlas Oğuz,
  Veselin Stoyanov, and Gargi Ghosh. 2021.
\newblock Multi-task retrieval for knowledge-intensive tasks.
\newblock In \emph{ACL/IJCNLP}.

\bibitem[{McCarthy(1959)}]{McCarthy_Programs59}
John McCarthy. 1959.
\newblock \href {http://www-formal.stanford.edu/jmc/mcc59.html} {Programs with
  common sense}.
\newblock In \emph{Proceedings of the {T}eddington Conference on the
  Mechanization of Thought Processes}, pages 75--91, London. Her Majesty's
  Stationary Office.

\bibitem[{Meng et~al.(2022)Meng, Bau, Andonian, and
  Belinkov}]{meng2022locating}
Kevin Meng, David Bau, Alex Andonian, and Yonatan Belinkov. 2022.
\newblock Locating and editing factual knowledge in gpt.
\newblock \emph{arXiv preprint arXiv:2202.05262}.

\bibitem[{Merrill et~al.(2021)Merrill, Goldberg, Schwartz, and
  Smith}]{Merrill2021ProvableLO}
William~Cooper Merrill, Yoav Goldberg, Roy Schwartz, and Noah~A. Smith. 2021.
\newblock Provable limitations of acquiring meaning from ungrounded form: What
  will future language models understand?
\newblock \emph{Transactions of the Association for Computational Linguistics},
  9:1047--1060.

\bibitem[{Metaxiotis et~al.(2002)Metaxiotis, Askounis, and
  Psarras}]{Metaxiotis2002ExpertSI}
Kostas~S. Metaxiotis, Dimitris Askounis, and John~E. Psarras. 2002.
\newblock Expert systems in production planning and scheduling: A
  state-of-the-art survey.
\newblock \emph{Journal of Intelligent Manufacturing}, 13:253--260.

\bibitem[{Michel et~al.(2019)Michel, Levy, and Neubig}]{michel2019sixteen}
Paul Michel, Omer Levy, and Graham Neubig. 2019.
\newblock \href {http://arxiv.org/abs/1905.10650} {Are sixteen heads really
  better than one?}

\bibitem[{Misra(2021)}]{Misra2021OnSC}
Kanishka Misra. 2021.
\newblock On semantic cognition, inductive generalization, and language models.
\newblock \emph{ArXiv}, abs/2111.02603.

\bibitem[{Mitchell et~al.(2021)Mitchell, Lin, Bosselut, Finn, and
  Manning}]{mitchell_fast_2021}
Eric Mitchell, Charles Lin, Antoine Bosselut, Chelsea Finn, and Christopher~D.
  Manning. 2021.
\newblock \href {http://arxiv.org/abs/2110.11309} {Fast {Model} {Editing} at
  {Scale}}.
\newblock \emph{arXiv:2110.11309 [cs]}.
\newblock ArXiv: 2110.11309.

\bibitem[{Mohankumar et~al.(2020)Mohankumar, Nema, Narasimhan, Khapra,
  Srinivasan, and Ravindran}]{mohankumar2020transparent}
Akash~Kumar Mohankumar, Preksha Nema, Sharan Narasimhan, Mitesh~M. Khapra,
  Balaji~Vasan Srinivasan, and Balaraman Ravindran. 2020.
\newblock \href {http://arxiv.org/abs/2004.14243} {Towards transparent and
  explainable attention models}.

\bibitem[{Narang et~al.(2020)Narang, Raffel, Lee, Roberts, Fiedel, and
  Malkan}]{Narang2020WT5TT}
Sharan Narang, Colin Raffel, Katherine Lee, Adam Roberts, Noah Fiedel, and
  Karishma Malkan. 2020.
\newblock Wt5?! training text-to-text models to explain their predictions.
\newblock \emph{ArXiv}, abs/2004.14546.

\bibitem[{Newell and Simon(1956)}]{NewellLogic56}
A.~Newell and H.~Simon. 1956.
\newblock \href {https://doi.org/10.1109/TIT.1956.1056797} {The logic theory
  machine--a complex information processing system}.
\newblock \emph{IRE Transactions on Information Theory}, 2(3):61--79.

\bibitem[{Nye et~al.(2021)Nye, Andreassen, Gur-Ari, Michalewski, Austin,
  Bieber, Dohan, Lewkowycz, Bosma, Luan, Sutton, and Odena}]{Nye2021ShowYW}
Maxwell Nye, Anders~Johan Andreassen, Guy Gur-Ari, Henryk Michalewski, Jacob
  Austin, David Bieber, David Dohan, Aitor Lewkowycz, Maarten Bosma, David
  Luan, Charles Sutton, and Augustus Odena. 2021.
\newblock Show your work: Scratchpads for intermediate computation with
  language models.
\newblock \emph{ArXiv}, abs/2112.00114.

\bibitem[{Och et~al.(2004)Och, Gildea, Khudanpur, Sarkar, Yamada, Fraser,
  Kumar, Shen, Smith, Eng, Jain, Jin, and Radev}]{och-etal-2004-smorgasbord}
Franz~Josef Och, Daniel Gildea, Sanjeev Khudanpur, Anoop Sarkar, Kenji Yamada,
  Alex Fraser, Shankar Kumar, Libin Shen, David Smith, Katherine Eng, Viren
  Jain, Zhen Jin, and Dragomir Radev. 2004.
\newblock \href {https://aclanthology.org/N04-1021} {A smorgasbord of features
  for statistical machine translation}.
\newblock In \emph{Proceedings of the Human Language Technology Conference of
  the North {A}merican Chapter of the Association for Computational
  Linguistics: {HLT}-{NAACL} 2004}, pages 161--168, Boston, Massachusetts, USA.
  Association for Computational Linguistics.

\bibitem[{Paranjape et~al.(2021)Paranjape, Michael, Ghazvininejad, Hajishirzi,
  and Zettlemoyer}]{paranjape-etal-2021-prompting}
Bhargavi Paranjape, Julian Michael, Marjan Ghazvininejad, Hannaneh Hajishirzi,
  and Luke Zettlemoyer. 2021.
\newblock \href {https://doi.org/10.18653/v1/2021.findings-acl.366} {Prompting
  contrastive explanations for commonsense reasoning tasks}.
\newblock In \emph{Findings of the Association for Computational Linguistics:
  ACL-IJCNLP 2021}, pages 4179--4192, Online. Association for Computational
  Linguistics.

\bibitem[{Peters et~al.(2019)Peters, Neumann, Logan, Schwartz, Joshi, Singh,
  and Smith}]{peters2019knowledge}
Matthew~E. Peters, Mark Neumann, Robert~L. Logan, Roy Schwartz, Vidur Joshi,
  Sameer Singh, and Noah~A. Smith. 2019.
\newblock \href {http://arxiv.org/abs/1909.04164} {Knowledge enhanced
  contextual word representations}.

\bibitem[{Peters et~al.(2018)Peters, Neumann, Zettlemoyer, and tau
  Yih}]{Peters2018DissectingCW}
Matthew~E. Peters, Mark Neumann, Luke Zettlemoyer, and Wen tau Yih. 2018.
\newblock Dissecting contextual word embeddings: Architecture and
  representation.
\newblock In \emph{EMNLP}.

\bibitem[{Petroni et~al.(2019)Petroni, Rocktäschel, Riedel, Lewis, Bakhtin,
  Wu, and Miller}]{petroni_language_2019}
Fabio Petroni, Tim Rocktäschel, Sebastian Riedel, Patrick Lewis, Anton
  Bakhtin, Yuxiang Wu, and Alexander Miller. 2019.
\newblock \href {https://doi.org/10.18653/v1/D19-1250} {Language {Models} as
  {Knowledge} {Bases}?}
\newblock In \emph{Proceedings of the 2019 {Conference} on {Empirical}
  {Methods} in {Natural} {Language} {Processing} and the 9th {International}
  {Joint} {Conference} on {Natural} {Language} {Processing}
  ({EMNLP}-{IJCNLP})}, pages 2463--2473, Hong Kong, China. Association for
  Computational Linguistics.

\bibitem[{Pezeshkpour et~al.(2021{\natexlab{a}})Pezeshkpour, Jain, Singh, and
  Wallace}]{pezeshkpour2021combining}
Pouya Pezeshkpour, Sarthak Jain, Sameer Singh, and Byron~C. Wallace.
  2021{\natexlab{a}}.
\newblock \href {http://arxiv.org/abs/2107.00323} {Combining feature and
  instance attribution to detect artifacts}.

\bibitem[{Pezeshkpour et~al.(2021{\natexlab{b}})Pezeshkpour, Jain, Wallace, and
  Singh}]{pezeshkpour2021empirical}
Pouya Pezeshkpour, Sarthak Jain, Byron~C. Wallace, and Sameer Singh.
  2021{\natexlab{b}}.
\newblock \href {http://arxiv.org/abs/2104.04128} {An empirical comparison of
  instance attribution methods for nlp}.

\bibitem[{Picado-Muiño(2011)}]{muino11}
David Picado-Muiño. 2011.
\newblock \href {https://doi.org/10.1016/j.ijar.2011.02.003} {Measuring and
  repairing inconsistency in probabilistic knowledge bases}.
\newblock \emph{Int. J. Approx. Reasoning}, 52:828--840.

\bibitem[{Poerner et~al.(2020)Poerner, Waltinger, and
  Sch{\"u}tze}]{poerner-etal-2020-e}
Nina Poerner, Ulli Waltinger, and Hinrich Sch{\"u}tze. 2020.
\newblock \href {https://doi.org/10.18653/v1/2020.findings-emnlp.71}
  {{E}-{BERT}: Efficient-yet-effective entity embeddings for {BERT}}.
\newblock In \emph{Findings of the Association for Computational Linguistics:
  EMNLP 2020}, pages 803--818, Online. Association for Computational
  Linguistics.

\bibitem[{Polu et~al.(2022)Polu, Han, Zheng, Baksys, Babuschkin, and
  Sutskever}]{Polu2022FormalMS}
Stanislas Polu, Jesse~Michael Han, Kunhao Zheng, Mantas Baksys, Igor
  Babuschkin, and Ilya Sutskever. 2022.
\newblock Formal mathematics statement curriculum learning.
\newblock \emph{ArXiv}, abs/2202.01344.

\bibitem[{Polu and Sutskever(2020)}]{Polu2020GenerativeLM}
Stanislas Polu and Ilya Sutskever. 2020.
\newblock Generative language modeling for automated theorem proving.
\newblock \emph{ArXiv}, abs/2009.03393.

\bibitem[{Qin and Eisner(2021)}]{qin2021learning}
Guanghui Qin and Jason Eisner. 2021.
\newblock \href {http://arxiv.org/abs/2104.06599} {Learning how to ask:
  Querying lms with mixtures of soft prompts}.

\bibitem[{Qin et~al.(2021)Qin, Lin, Takanobu, Liu, Li, Ji, Huang, Sun, and
  Zhou}]{Qin2021ERICAIE}
Yujia Qin, Yankai Lin, Ryuichi Takanobu, Zhiyuan Liu, Peng Li, Heng Ji, Minlie
  Huang, Maosong Sun, and Jie Zhou. 2021.
\newblock Erica: Improving entity and relation understanding for pre-trained
  language models via contrastive learning.
\newblock In \emph{ACL/IJCNLP}.

\bibitem[{Radford and Narasimhan(2018)}]{Radford2018ImprovingLU}
Alec Radford and Karthik Narasimhan. 2018.
\newblock Improving language understanding by generative pre-training.

\bibitem[{Radford et~al.(2019)Radford, Wu, Child, Luan, Amodei, and
  Sutskever}]{Radford2019LanguageMA}
Alec Radford, Jeff Wu, Rewon Child, David Luan, Dario Amodei, and Ilya
  Sutskever. 2019.
\newblock Language models are unsupervised multitask learners.

\bibitem[{Raffel et~al.(2020)Raffel, Shazeer, Roberts, Lee, Narang, Matena,
  Zhou, Li, and Liu}]{Raffel2020ExploringTL}
Colin Raffel, Noam~M. Shazeer, Adam Roberts, Katherine Lee, Sharan Narang,
  Michael Matena, Yanqi Zhou, Wei Li, and Peter~J. Liu. 2020.
\newblock Exploring the limits of transfer learning with a unified text-to-text
  transformer.
\newblock \emph{ArXiv}, abs/1910.10683.

\bibitem[{Ravichander et~al.(2020)Ravichander, Hovy, Suleman, Trischler, and
  Cheung}]{ravichander-etal-2020-systematicity}
Abhilasha Ravichander, Eduard Hovy, Kaheer Suleman, Adam Trischler, and Jackie
  Chi~Kit Cheung. 2020.
\newblock \href {https://aclanthology.org/2020.starsem-1.10} {On the
  systematicity of probing contextualized word representations: The case of
  hypernymy in {BERT}}.
\newblock In \emph{Proceedings of the Ninth Joint Conference on Lexical and
  Computational Semantics}, pages 88--102, Barcelona, Spain (Online).
  Association for Computational Linguistics.

\bibitem[{Ri et~al.(2021)Ri, Yamada, and Tsuruoka}]{Ri2021mLUKETP}
Ryokan Ri, Ikuya Yamada, and Yoshimasa Tsuruoka. 2021.
\newblock mluke: The power of entity representations in multilingual pretrained
  language models.
\newblock \emph{ArXiv}, abs/2110.08151.

\bibitem[{Rogers et~al.(2020)Rogers, Kovaleva, and
  Rumshisky}]{rogers_primer_2020}
Anna Rogers, Olga Kovaleva, and Anna Rumshisky. 2020.
\newblock \href {http://arxiv.org/abs/2002.12327} {A {Primer} in {BERTology}:
  {What} we know about how {BERT} works}.
\newblock \emph{arXiv:2002.12327 [cs]}.
\newblock ArXiv: 2002.12327.

\bibitem[{Rosset et~al.(2020)Rosset, Xiong, Phan, Song, Bennett, and
  Tiwary}]{Rosset2020KnowledgeAwareLM}
Corby Rosset, Chenyan Xiong, Minh~Hieu Phan, Xia Song, Paul Bennett, and
  Saurabh Tiwary. 2020.
\newblock Knowledge-aware language model pretraining.
\newblock \emph{ArXiv}, abs/2007.00655.

\bibitem[{Safavi and Koutra(2021)}]{safavi2021relational}
Tara Safavi and Danai Koutra. 2021.
\newblock \href {http://arxiv.org/abs/2104.05837} {Relational world knowledge
  representation in contextual language models: A review}.

\bibitem[{Saha et~al.(2020)Saha, Ghosh, Srivastava, and
  Bansal}]{Saha2020PRoverPG}
Swarnadeep Saha, Sayan Ghosh, Shashank Srivastava, and Mohit Bansal. 2020.
\newblock Prover: Proof generation for interpretable reasoning over rules.
\newblock \emph{ArXiv}, abs/2010.02830.

\bibitem[{Sap et~al.(2019)Sap, Le~Bras, Allaway, Bhagavatula, Lourie, Rashkin,
  Roof, Smith, and Choi}]{sap2019atomic}
Maarten Sap, Ronan Le~Bras, Emily Allaway, Chandra Bhagavatula, Nicholas
  Lourie, Hannah Rashkin, Brendan Roof, Noah~A Smith, and Yejin Choi. 2019.
\newblock Atomic: An atlas of machine commonsense for if-then reasoning.
\newblock In \emph{Proceedings of the AAAI Conference on Artificial
  Intelligence}, volume~33, pages 3027--3035.

\bibitem[{Saxton et~al.(2019)Saxton, Grefenstette, Hill, and
  Kohli}]{Saxton2019AnalysingMR}
David Saxton, Edward Grefenstette, Felix Hill, and Pushmeet Kohli. 2019.
\newblock Analysing mathematical reasoning abilities of neural models.
\newblock \emph{ArXiv}, abs/1904.01557.

\bibitem[{Schick and Schütze(2021{\natexlab{a}})}]{schick2021exploiting}
Timo Schick and Hinrich Schütze. 2021{\natexlab{a}}.
\newblock \href {http://arxiv.org/abs/2001.07676} {Exploiting cloze questions
  for few shot text classification and natural language inference}.

\bibitem[{Schick and Schütze(2021{\natexlab{b}})}]{schick2021its}
Timo Schick and Hinrich Schütze. 2021{\natexlab{b}}.
\newblock \href {http://arxiv.org/abs/2009.07118} {It's not just size that
  matters: Small language models are also few-shot learners}.

\bibitem[{Schramowski et~al.(2021)Schramowski, Friedrich, Tauchmann, and
  Kersting}]{Schramowski21}
Patrick Schramowski, Felix Friedrich, Christopher Tauchmann, and Kristian
  Kersting. 2021.
\newblock \href {http://arxiv.org/abs/2110.02058} {Interactively generating
  explanations for transformer language models}.
\newblock \emph{CoRR}, abs/2110.02058.

\bibitem[{Sennrich et~al.(2016)Sennrich, Haddow, and
  Birch}]{sennrich-etal-2016-improving}
Rico Sennrich, Barry Haddow, and Alexandra Birch. 2016.
\newblock \href {https://doi.org/10.18653/v1/P16-1009} {Improving neural
  machine translation models with monolingual data}.
\newblock In \emph{Proceedings of the 54th Annual Meeting of the Association
  for Computational Linguistics (Volume 1: Long Papers)}, pages 86--96, Berlin,
  Germany. Association for Computational Linguistics.

\bibitem[{Serrano and Smith(2019)}]{serrano2019attention}
Sofia Serrano and Noah~A. Smith. 2019.
\newblock \href {http://arxiv.org/abs/1906.03731} {Is attention interpretable?}

\bibitem[{Shin et~al.(2020)Shin, Razeghi, Logan, Wallace, and
  Singh}]{shin2020autoprompt}
Taylor Shin, Yasaman Razeghi, Robert~L. Logan, Eric Wallace, and Sameer Singh.
  2020.
\newblock \href {http://arxiv.org/abs/2010.15980} {Autoprompt: Eliciting
  knowledge from language models with automatically generated prompts}.

\bibitem[{Sun et~al.(2021)Sun, Verga, Dhingra, Salakhutdinov, and
  Cohen}]{sun_opql}
Haitian Sun, Pat Verga, Bhuwan Dhingra, Ruslan Salakhutdinov, and William~W.
  Cohen. 2021.
\newblock \href {http://arxiv.org/abs/2102.07043} {Reasoning over virtual
  knowledge bases with open predicate relations}.
\newblock \emph{CoRR}, abs/2102.07043.

\bibitem[{Sun et~al.(2019)Sun, Wang, Li, Feng, Chen, Zhang, Tian, Zhu, Tian,
  and Wu}]{Sun2019ERNIEER}
Yu~Sun, Shuohuan Wang, Yukun Li, Shikun Feng, Xuyi Chen, Han Zhang, Xin Tian,
  Danxiang Zhu, Hao Tian, and Hua Wu. 2019.
\newblock Ernie: Enhanced representation through knowledge integration.
\newblock \emph{ArXiv}, abs/1904.09223.

\bibitem[{Sung et~al.(2021)Sung, Lee, Yi, Jeon, Kim, and
  Kang}]{sung2021language}
Mujeen Sung, Jinhyuk Lee, Sean Yi, Minji Jeon, Sungdong Kim, and Jaewoo Kang.
  2021.
\newblock \href {http://arxiv.org/abs/2109.07154} {Can language models be
  biomedical knowledge bases?}

\bibitem[{Tafjord et~al.(2021)Tafjord, Dalvi, and
  Clark}]{tafjord-etal-2021-proofwriter}
Oyvind Tafjord, Bhavana Dalvi, and Peter Clark. 2021.
\newblock \href {https://doi.org/10.18653/v1/2021.findings-acl.317}
  {{P}roof{W}riter: Generating implications, proofs, and abductive statements
  over natural language}.
\newblock In \emph{Findings of the Association for Computational Linguistics:
  ACL-IJCNLP 2021}, pages 3621--3634, Online. Association for Computational
  Linguistics.

\bibitem[{Talmor et~al.(2020{\natexlab{a}})Talmor, Elazar, Goldberg, and
  Berant}]{talmor_olmpics_2020}
Alon Talmor, Yanai Elazar, Yoav Goldberg, and Jonathan Berant.
  2020{\natexlab{a}}.
\newblock \href {http://arxiv.org/abs/1912.13283} {{oLMpics} -- {On} what
  {Language} {Model} {Pre}-training {Captures}}.
\newblock \emph{arXiv:1912.13283 [cs]}.
\newblock ArXiv: 1912.13283.

\bibitem[{Talmor et~al.(2019)Talmor, Herzig, Lourie, and
  Berant}]{talmor-etal-2019-commonsenseqa}
Alon Talmor, Jonathan Herzig, Nicholas Lourie, and Jonathan Berant. 2019.
\newblock \href {https://doi.org/10.18653/v1/N19-1421} {{C}ommonsense{QA}: A
  question answering challenge targeting commonsense knowledge}.
\newblock In \emph{Proceedings of the 2019 Conference of the North {A}merican
  Chapter of the Association for Computational Linguistics: Human Language
  Technologies, Volume 1 (Long and Short Papers)}, pages 4149--4158,
  Minneapolis, Minnesota. Association for Computational Linguistics.

\bibitem[{Talmor et~al.(2020{\natexlab{b}})Talmor, Tafjord, Clark, Goldberg,
  and Berant}]{talmor_leap--thought_2020}
Alon Talmor, Oyvind Tafjord, Peter Clark, Yoav Goldberg, and Jonathan Berant.
  2020{\natexlab{b}}.
\newblock \href {http://arxiv.org/abs/2006.06609} {Leap-{Of}-{Thought}:
  {Teaching} {Pre}-{Trained} {Models} to {Systematically} {Reason} {Over}
  {Implicit} {Knowledge}}.
\newblock \emph{arXiv:2006.06609 [cs]}.
\newblock ArXiv: 2006.06609.

\bibitem[{Taylor(1953)}]{taylor1953cloze}
Wilson~L Taylor. 1953.
\newblock “cloze procedure”: A new tool for measuring readability.
\newblock \emph{Journalism quarterly}, 30(4):415--433.

\bibitem[{Tenney et~al.(2019{\natexlab{a}})Tenney, Das, and
  Pavlick}]{tenney2019bert}
Ian Tenney, Dipanjan Das, and Ellie Pavlick. 2019{\natexlab{a}}.
\newblock \href {http://arxiv.org/abs/1905.05950} {Bert rediscovers the
  classical nlp pipeline}.

\bibitem[{Tenney et~al.(2019{\natexlab{b}})Tenney, Xia, Chen, Wang, Poliak,
  McCoy, Kim, Durme, Bowman, Das, and Pavlick}]{tenney2018what}
Ian Tenney, Patrick Xia, Berlin Chen, Alex Wang, Adam Poliak, R~Thomas McCoy,
  Najoung Kim, Benjamin~Van Durme, Sam Bowman, Dipanjan Das, and Ellie Pavlick.
  2019{\natexlab{b}}.
\newblock \href {https://openreview.net/forum?id=SJzSgnRcKX} {What do you learn
  from context? probing for sentence structure in contextualized word
  representations}.
\newblock In \emph{International Conference on Learning Representations}.

\bibitem[{Vaswani et~al.(2017)Vaswani, Shazeer, Parmar, Uszkoreit, Jones,
  Gomez, Kaiser, and Polosukhin}]{vaswani17attention}
Ashish Vaswani, Noam Shazeer, Niki Parmar, Jakob Uszkoreit, Llion Jones,
  Aidan~N. Gomez, undefinedukasz Kaiser, and Illia Polosukhin. 2017.
\newblock Attention is all you need.
\newblock In \emph{Proceedings of the 31st International Conference on Neural
  Information Processing Systems}, NIPS'17, page 6000–6010, Red Hook, NY,
  USA. Curran Associates Inc.

\bibitem[{Verga et~al.(2020)Verga, Sun, Soares, and Cohen}]{facts-as-experts}
Pat Verga, Haitian Sun, Livio~Baldini Soares, and William~W. Cohen. 2020.
\newblock \href {http://arxiv.org/abs/2007.00849} {Facts as experts: Adaptable
  and interpretable neural memory over symbolic knowledge}.
\newblock \emph{CoRR}, abs/2007.00849.

\bibitem[{Vig and Belinkov(2019)}]{vig2019analyzing}
Jesse Vig and Yonatan Belinkov. 2019.
\newblock \href {http://arxiv.org/abs/1906.04284} {Analyzing the structure of
  attention in a transformer language model}.

\bibitem[{Vrande\v{c}i\'{c} and Kr\"{o}tzsch(2014)}]{wikidata}
Denny Vrande\v{c}i\'{c} and Markus Kr\"{o}tzsch. 2014.
\newblock \href {https://doi.org/10.1145/2629489} {Wikidata: A free
  collaborative knowledgebase}.
\newblock \emph{Commun. ACM}, 57(10):78–85.

\bibitem[{Wallat et~al.(2021)Wallat, Singh, and Anand}]{wallat_bertnesia_2021}
Jonas Wallat, Jaspreet Singh, and Avishek Anand. 2021.
\newblock \href {http://arxiv.org/abs/2106.02902} {{BERTnesia}: {Investigating}
  the capture and forgetting of knowledge in {BERT}}.
\newblock \emph{arXiv:2106.02902 [cs]}.
\newblock ArXiv: 2106.02902.

\bibitem[{Wang et~al.(2020)Wang, Liu, and Song}]{wang_language_2020}
Chenguang Wang, Xiao Liu, and Dawn Song. 2020.
\newblock \href {http://arxiv.org/abs/2010.11967} {Language {Models} are {Open}
  {Knowledge} {Graphs}}.
\newblock \emph{arXiv:2010.11967 [cs]}.
\newblock ArXiv: 2010.11967.

\bibitem[{Wang et~al.(2021{\natexlab{a}})Wang, Fang, Khabsa, Mao, and
  Ma}]{wang2021entailment}
Sinong Wang, Han Fang, Madian Khabsa, Hanzi Mao, and Hao Ma.
  2021{\natexlab{a}}.
\newblock \href {http://arxiv.org/abs/2104.14690} {Entailment as few-shot
  learner}.

\bibitem[{Wang et~al.(2021{\natexlab{b}})Wang, Gao, Zhu, Liu, Li, and
  Tang}]{Wang2021KEPLERAU}
Xiaozhi Wang, Tianyu Gao, Zhaocheng Zhu, Zhiyuan Liu, Juan-Zi Li, and Jian
  Tang. 2021{\natexlab{b}}.
\newblock Kepler: A unified model for knowledge embedding and pre-trained
  language representation.
\newblock \emph{Transactions of the Association for Computational Linguistics},
  9:176--194.

\bibitem[{Wang et~al.(2021{\natexlab{c}})Wang, Gao, Zhu, Zhang, Liu, Li, and
  Tang}]{wang2021}
Xiaozhi Wang, Tianyu Gao, Zhaocheng Zhu, Zhengyan Zhang, Zhiyuan Liu, Juanzi
  Li, and Jian Tang. 2021{\natexlab{c}}.
\newblock \href {https://doi.org/10.1162/tacl_a_00360} {{KEPLER: A Unified
  Model for Knowledge Embedding and Pre-trained Language Representation}}.
\newblock \emph{Transactions of the Association for Computational Linguistics},
  9:176--194.

\bibitem[{Wei et~al.(2022)Wei, Wang, Schuurmans, Bosma, Chi, Le, and
  Zhou}]{Wei2022ChainOT}
Jason Wei, Xuezhi Wang, Dale Schuurmans, Maarten Bosma, Ed~Chi, Quoc Le, and
  Denny Zhou. 2022.
\newblock Chain of thought prompting elicits reasoning in large language
  models.
\newblock \emph{ArXiv}, abs/2201.11903.

\bibitem[{Wei et~al.(2021)Wei, Wang, Zhang, Bhatia, and
  Arnold}]{wei2021knowledge}
Xiaokai Wei, Shen Wang, Dejiao Zhang, Parminder Bhatia, and Andrew Arnold.
  2021.
\newblock \href {http://arxiv.org/abs/2110.08455} {Knowledge enhanced
  pretrained language models: A compreshensive survey}.

\bibitem[{West et~al.(2021)West, Bhagavatula, Hessel, Hwang, Jiang, Bras, Lu,
  Welleck, and Choi}]{west_symbolic_2021}
Peter West, Chandra Bhagavatula, Jack Hessel, Jena~D. Hwang, Liwei Jiang,
  Ronan~Le Bras, Ximing Lu, Sean Welleck, and Yejin Choi. 2021.
\newblock \href {http://arxiv.org/abs/2110.07178} {Symbolic {Knowledge}
  {Distillation}: from {General} {Language} {Models} to {Commonsense}
  {Models}}.
\newblock \emph{arXiv:2110.07178 [cs]}.
\newblock ArXiv: 2110.07178.

\bibitem[{Weston et~al.(2015)Weston, Chopra, and Bordes}]{weston2015memory}
Jason Weston, Sumit Chopra, and Antoine Bordes. 2015.
\newblock \href {http://arxiv.org/abs/1410.3916} {Memory networks}.

\bibitem[{Wiegreffe and Pinter(2019)}]{wiegreffe2019attention}
Sarah Wiegreffe and Yuval Pinter. 2019.
\newblock \href {http://arxiv.org/abs/1908.04626} {Attention is not not
  explanation}.

\bibitem[{Wieting and Gimpel(2018)}]{wieting-gimpel-2018-paranmt}
John Wieting and Kevin Gimpel. 2018.
\newblock \href {https://doi.org/10.18653/v1/P18-1042} {{P}ara{NMT}-50{M}:
  Pushing the limits of paraphrastic sentence embeddings with millions of
  machine translations}.
\newblock In \emph{Proceedings of the 56th Annual Meeting of the Association
  for Computational Linguistics (Volume 1: Long Papers)}, pages 451--462,
  Melbourne, Australia. Association for Computational Linguistics.

\bibitem[{Yamada et~al.(2020)Yamada, Asai, Shindo, Takeda, and
  Matsumoto}]{Yamada2020LUKEDC}
Ikuya Yamada, Akari Asai, Hiroyuki Shindo, Hideaki Takeda, and Yuji Matsumoto.
  2020.
\newblock Luke: Deep contextualized entity representations with entity-aware
  self-attention.
\newblock In \emph{EMNLP}.

\bibitem[{Yao et~al.(2019)Yao, Mao, and Luo}]{Yao2019KGBERTBF}
Liang Yao, Chengsheng Mao, and Yuan Luo. 2019.
\newblock Kg-bert: Bert for knowledge graph completion.
\newblock \emph{ArXiv}, abs/1909.03193.

\bibitem[{Yasunaga et~al.(2021)Yasunaga, Ren, Bosselut, Liang, and
  Leskovec}]{yasunaga2021qagnn}
Michihiro Yasunaga, Hongyu Ren, Antoine Bosselut, Percy Liang, and Jure
  Leskovec. 2021.
\newblock \href {http://arxiv.org/abs/2104.06378} {Qa-gnn: Reasoning with
  language models and knowledge graphs for question answering}.

\bibitem[{Yogatama et~al.(2021)Yogatama, d'Autume, and
  Kong}]{yogatama_adaptive_2021}
Dani Yogatama, Cyprien de~Masson d'Autume, and Lingpeng Kong. 2021.
\newblock \href {http://arxiv.org/abs/2102.02557} {Adaptive {Semiparametric}
  {Language} {Models}}.
\newblock \emph{arXiv:2102.02557 [cs]}.
\newblock ArXiv: 2102.02557.

\bibitem[{Zhang et~al.(2021)Zhang, Li, Chen, Deng, Bi, Tan, Huang, and
  Chen}]{Zhang2021DifferentiablePM}
Ningyu Zhang, Luoqiu Li, Xiang Chen, Shumin Deng, Zhen Bi, Chuanqi Tan, Fei
  Huang, and Huajun Chen. 2021.
\newblock Differentiable prompt makes pre-trained language models better
  few-shot learners.
\newblock \emph{ArXiv}, abs/2108.13161.

\bibitem[{Zhang and Nivre(2011)}]{zhang-nivre-2011-transition}
Yue Zhang and Joakim Nivre. 2011.
\newblock \href {https://aclanthology.org/P11-2033} {Transition-based
  dependency parsing with rich non-local features}.
\newblock In \emph{Proceedings of the 49th Annual Meeting of the Association
  for Computational Linguistics: Human Language Technologies}, pages 188--193,
  Portland, Oregon, USA. Association for Computational Linguistics.

\bibitem[{Zhang et~al.(2019)Zhang, Han, Liu, Jiang, Sun, and
  Liu}]{zhang-ernie-2019}
Zhengyan Zhang, Xu~Han, Zhiyuan Liu, Xin Jiang, Maosong Sun, and Qun Liu. 2019.
\newblock \href {http://arxiv.org/abs/1905.07129} {{ERNIE:} enhanced language
  representation with informative entities}.
\newblock \emph{CoRR}, abs/1905.07129.

\bibitem[{Zhao et~al.(2021)Zhao, Pascual, Brunner, and
  Wattenhofer}]{zhao2021nonlinearity}
Sumu Zhao, Damian Pascual, Gino Brunner, and Roger Wattenhofer. 2021.
\newblock \href {http://arxiv.org/abs/2101.04547} {Of non-linearity and
  commutativity in bert}.

\bibitem[{Zhong et~al.(2021)Zhong, Friedman, and Chen}]{zhong2021factual}
Zexuan Zhong, Dan Friedman, and Danqi Chen. 2021.
\newblock \href {http://arxiv.org/abs/2104.05240} {Factual probing is [mask]:
  Learning vs. learning to recall}.

\bibitem[{Zhou et~al.(2020)Zhou, Zhang, Cui, and
  Huang}]{Zhou_Zhang_Cui_Huang_2020}
Xuhui Zhou, Yue Zhang, Leyang Cui, and Dandan Huang. 2020.
\newblock \href {https://doi.org/10.1609/aaai.v34i05.6523} {Evaluating
  commonsense in pre-trained language models}.
\newblock \emph{Proceedings of the AAAI Conference on Artificial Intelligence},
  34(05):9733--9740.

\bibitem[{Zhu et~al.(2020)Zhu, Rawat, Zaheer, Bhojanapalli, Li, Yu, and
  Kumar}]{zhu_modifying_2020}
Chen Zhu, Ankit~Singh Rawat, Manzil Zaheer, Srinadh Bhojanapalli, Daliang Li,
  Felix Yu, and Sanjiv Kumar. 2020.
\newblock \href {http://arxiv.org/abs/2012.00363} {Modifying {Memories} in
  {Transformer} {Models}}.
\newblock \emph{arXiv:2012.00363 [cs]}.
\newblock ArXiv: 2012.00363.

\bibitem[{Zylberajch et~al.(2021)Zylberajch, Lertvittayakumjorn, and
  Toni}]{zylberajch-etal-2021-hildif}
Hugo Zylberajch, Piyawat Lertvittayakumjorn, and Francesca Toni. 2021.
\newblock \href {https://doi.org/10.18653/v1/2021.internlp-1.1} {{HILDIF}:
  {I}nteractive debugging of {NLI} models using influence functions}.
\newblock In \emph{Proceedings of the First Workshop on Interactive Learning
  for Natural Language Processing}, pages 1--6, Online. Association for
  Computational Linguistics.

\end{thebibliography}
\bibliographystyle{acl_natbib}

\section*{Appendix}
\section{Models}

\begin{table*}[]
\begin{adjustbox}{width=1\textwidth,center}
\begin{tabular}{llccccc}
\toprule
\textbf{Name}            & \textbf{Author}          & \textbf{External KG} & \textbf{Non-Parametric Memory} & \textbf{Pretraining} & \textbf{Attention} & \textbf{Retrieval} \\ \midrule
QA-GNN        & \citet{yasunaga2021qagnn}   & \checkmark             &                       &             &           &           \\ \hline
KnowBERT        & \citet{peters2019knowledge}    &             &   \checkmark     &  \checkmark           &           &           \\ \hline
KGLM            & \citet{logan2019baracks}   & \checkmark            &                       &             &           &           \\ \hline
NKLM            & \citet{ahn2017neural}        &             &             \checkmark &            &           &           \\ \hline
REALM           & \citet{guu_realm_2020}        &             &                       &             &           & \checkmark           \\ \hline
EAE             & \citet{entities-as-experts}      &             &     \checkmark    &           &           &           \\ \hline
FAE             & \citet{facts-as-experts}      &             &     \checkmark       &   \checkmark            &           &           \\ \hline
OPQL-LM         & \citet{sun_opql}        &             &     \checkmark    & \checkmark              &           &           \\ \hline
TOME            & \citet{dejong2021mention}    &             &     \checkmark        & \checkmark              &  \checkmark         &           \\ \hline
RAG             & \citet{lewis_retrieval-augmented_2021}      &             &    \checkmark                     &             &           &   \checkmark  \\ \hline
LUKE            & \citet{Yamada2020LUKEDC}     &             &                       & \checkmark            &     \checkmark      &           \\ \hline
mLUKE           & \citet{Ri2021mLUKETP}         &             &                       &   \checkmark          &    \checkmark       &           \\ \hline
ERNIE           & \citet{zhang-ernie-2019}      &             &  \checkmark                     &  \checkmark           &           &           \\ \hline
ERICA           & \citet{Qin2021ERICAIE}        &             &                       &  \checkmark           &           &           \\ \hline
KEPLER          & \citet{wang2021}       &             &                       & \checkmark            &      \checkmark     &           \\ \hline
SPALM           & \citet{yogatama_adaptive_2021}   &             &   \checkmark                    &             &           &           \\ \hline
GRF             & \citet{ji_survey_2021}         & \checkmark            &                       &             &           &           \\ \hline
RelationLM      & \citet{Liu2022RelationalMA}.        &             &   \checkmark                    &             &           &           \\ \hline
HTLM            & \citet{aghajanyan_htlm_2021} &             &                       &  \checkmark           &           &           \\ \hline
CM3             & \citet{Aghajanyan2022CM3AC} &             &                       &  \checkmark           &           &           \\ \hline
K-BERT          & \citet{Liu2020KBERTEL}        &             &   \checkmark                    &             &           &           \\ \hline
ERNIE           & \citet{Sun2019ERNIEER}        &             &                       &  \checkmark           &           &           \\ \hline
KG-BERT         & \citet{Yao2019KGBERTBF}        &             &                       &  \checkmark           &           &           \\ \hline
BERT-MK         & \citet{he-etal-2020-bert}         &   \checkmark          &                       &             &           &           \\ \hline
KALM            & \citet{Rosset2020KnowledgeAwareLM}     &             &    \checkmark                   & \checkmark            &           &           \\ \hline
DrKIT           & \citet{dhingra_time-aware_2021}    &             &                       & \checkmark            &           &           \\ \hline
MBPA++          & \citet{dAutume2019EpisodicMI}   &             &   \checkmark                    &             &           &           \\ \hline
Multitask Model & \citet{Maillard2021MultitaskRF}   &             &                       &             &           &   \checkmark        \\ \bottomrule
\end{tabular}
\end{adjustbox}
\caption{Models and the associated modifications to improve knowledge within their parameters.}
\end{table*}

In the main sections, we detail more general aspects of LMs. To follow the sections, we cover the different models that fall under the \lmkb paradigm and the solutions proposed by these models that improve LM performance.

These models often incorporate knowledge explicitly through a combination of several means: via some form of \textbf{pretraining} strategy that explicitly encodes entity-level or relation-level data, via the integration of \textbf{external memory} to an existing LM, via an \textbf{attention-based mechanism}, or via a \textbf{retrieval-based} model that gathers the appropriate nodes of a KG. Models that implicitly contain knowledge, such as existing pretrained LMs, are not covered in this section, as we focus on explicit incorporation of knowledge.

\subsection{External Memory} Existing LMs are often unable to store localized information about facts and specific knowledge; on the other hand, KBs have been known to store information about millions of entities in an interpretable fashion. Knowing this, existing research focus on taking advantage of the storage capabilities of KBs and integrating this capability into LMs \cite{wei2021knowledge}.

\citet{heinzerling_language_2021} introduce initial work understanding how \lmkb can be used in more general settings, especially in integrating knowledge about general and rare entities, since representing millions of entities within LMs is difficult when LMs have a limited vocabulary. Recent work often forgo the potential to deploy these \lmkb in the real world. To improve existing LM and overcome memory limitations, recent work focuses on the goal of improving memory for LMs to store more information about external information, such as information about entities in a sentence.

External memory has previously been used for neural networks \cite{Bahdanau2015NeuralMT, weston2015memory, graves2014neural, ahn2017neural}. \citet{ahn2017neural} specifically propose the \newterm{NKLM} to take advantage of external knowledge and exploit factual knowledge. Now, a large number of works now integrate external memory to improve performance on knowledge-intensive tasks. To summarize these works, we focus on detailing the different strategies of integrating external memory within LMs and how they can be further improved on.

\subsubsection{External Knowledge Graphs}
\label{app:kgs}



Prior work on \lmkb involves the explicit creation of external knowledge graphs (KG) that are incorporated into models, such as in the model \newterm{KGLM} \cite{logan2019baracks}. KGLM looks at explicitly incorporating KGs into LMs. KGLM can access facts which are stored in a KG, growing constantly as new facts are added. The model selectively grows the KG if a fact is not there or refers back to the KG to pick an existing fact.

While not explicitly a model, \newterm{GRF} \cite{ji2020language} propose a method to enable multi-hop reasoning with transformers. To do so, the model first encodes multi-relational graphs to obtain representations for concepts, then reasons over multiple relational paths to generate the concept, and finally chooses the output by determining the probability of obtaining the concept from the KG versus choosing a word from its innate vocabulary.

Other work also evaluate other domains. \citet{he-etal-2020-bert} introduce \newterm{BERT-MK}, which integrates contextualized knowledge from a medical KG, which shows improvement over existing biomedical models on entity typing and relation classification tasks.



\subsubsection{Non-parametric Memory}
Non-parametric memory can be seen as memory that is used in addition to internal LM memory. Existing work integrates non-parametric memory with LMs to improve performance on downstream tasks.

\citet{ahn2017neural} adapts an early version of non-parametric memory with neural models using \newterm{NKLM}. NKLM is able to combine
symbolic knowledge provided from a KG with an RNN \cite{hochreiter1997long}.

Other work adopt a similar strategy via embeddings to incorporate entity knowledge. Early work on \newterm{KnowBERT} \cite{peters2019knowledge} introduces entity vectors that are computed from mention-span representations that are obtained from BERT \cite{devlin-etal-2019-bert} to form entity-span representations. Other work follow similarly: \citet{Fvry2020EntitiesAE} introduce \newterm{EAE}, a LM augmented with entity memory to keep track of facts about entities. Building off of this, \newterm{FAE} \cite{facts-as-experts} adds an additional memory that encodes triples from a symbolic KB that can be accessed with key-value memory to extract facts and improve predictions. Other models such as \newterm{ERNIE}, \newterm{KEPLER}, and \newterm{KALM} use existing algorithms to embed entity and relation knowledge into embeddings \cite{zhang-ernie-2019, Wang2021KEPLERAU, Rosset2020KnowledgeAwareLM}. \newterm{RAG} uses vector indices of Wikipedia to access latent information during inference on knowledge-intensive tasks \cite{lewis_retrieval-augmented_2021}

Introduced by \cite{dhingra2020differentiable},  \newterm{DrKIT} includes a virtual knowledge base (VKB), which is a "soft knowledge base" that is used in conjunction with neural methods to compensate for structure and find answers to questions. \citet{sun_opql} improve the idea of a VKB and introduces an additional strategy to learn entity and information through \newterm{OPQL}, utilizing key-value memory to learn relationships between entities and relations. \citet{dejong2021mention} take the idea of a VKB and insert it into \newterm{TOME} as non-parametric memory to improve reasoning over various knowledge sources.

Episodic non-parametric memory can also be introduced to LMs to remember specific knowledge, including both short-term and long-term contexts \cite{dAutume2019EpisodicMI, yogatama_adaptive_2021, Liu2022RelationalMA}.

\citet{Liu2020KBERTEL} try a different strategy by introducing a visible matrix for their model \newterm{K-BERT} to control the impact of certain knowledge injected.

\subsubsection{Attention over Memory}
Attention mechanisms over specific types of memory can extract salient information during inference. Early iterations of attention over memory can be seen in KnowBERT \cite{peters2019knowledge} where they introduce \newterm{KAR}, a form of multi-headed attention between word representations and knowledge-enchanced entity span vectors. On the other hand, in TOME \cite{dejong2021mention}, they employ attention over an entire VKB within a transformer model to retrieve the relevant pieces of knowledge.

Other work modify existing transformer layers to add an additional self-attention mechanisms over entity knowledge, specifically looking at using the attention to identify the query mechanisms required depending on the attending token and token attended to \cite{Yamada2020LUKEDC, Ri2021mLUKETP}.

\subsection{Pretraining}
Models have been known to encode knowledge within their parameters through unsupervised methods during pretraining. For example, given the example \texttt{Obama is from [MASK]}, using a masked language modeling (MLM) objective, we would expect the model to predict \texttt{Hawaii}. However, there is no explicit integration of any entity-level or relation-level knowledge through this objective. As a result, a number of work have sought to incorporate entity-level or relation-level knowledge through pretraining strategies and loss function modifications.

Various models focus on different pretraining strategies, whether that be through the augmentation of the input or modification of loss functions or others. FAE adapts from the the EAE model to introduce a pretraining scheme modeled as a cloze-type Question Answering (QA) task \cite{facts-as-experts}. Other pretraining tasks include an augmentation of the existing masked language modeling (MLM) task by masking the entities and predicting the entities during pretraining \cite{sun_opql, Yamada2020LUKEDC, Ri2021mLUKETP, zhang-ernie-2019} or introducing multiple preexisting pretraining tasks to be used in conjunction with each other, such as MLM \cite{Yamada2020LUKEDC, Ri2021mLUKETP}. \citet{Sun2019ERNIEER} introduce three levels of masking for their pretraining task: basic-level, phrase-level, and entity-level masking. Other work introduce a special loss function in conjunction with the standard MLM objective that focuses on predicting entities \cite{Rosset2020KnowledgeAwareLM}, linking entities to text \cite{entities-as-experts, facts-as-experts, sun_opql, dejong2021mention}, objectives focusing on semantic understanding of relations and entities \cite{Qin2021ERICAIE}, or knowledge embedding objective where entities and relations are encoded in KGs as distributed representations KE \cite{wang_language_2020}.

Some models are trained on inputs augmented specifically to encode better representations of entities and relations. For example, \citet{dejong2021mention} inserts start and end entity tokens around each entity in the input. Others mask both relations and entities during pretraining, like in \citet{sun_opql}. \citet{Yao2019KGBERTBF} also imbues entity and relation information by taking the entities and relations in a sequence as input into existing LMs. 

\citet{aghajanyan_htlm_2021, Aghajanyan2022CM3AC} look at methods to incorporate more knowledge with data not commonly used in unimodal and multimodal settings. In \citet{aghajanyan_htlm_2021, Aghajanyan2022CM3AC}, both \newterm{HTLM} and \newterm{CM3} models apply scraped HTML and find that pretraining with size hints and prompting with BART \cite{lewis-etal-2020-bart} can creative effective transfer to a wide range of downstream tasks and supervision levels. These approaches show the potential for web-scrapped data to be used as viable signals for model pretraining on knowledge-intensive tasks.

\subsection{Retrieval}
Some models can capture more knowledge in modular and interpretable ways via retrieval methods. \citet{guu_realm_2020} introduce \newterm{REALM} which contain ways to extract knowledge by applying retrieval-based methods during pretraining, finetuning, and inference. Their method allows retrieval of Wikipedia knowledge and ability for the model to decide what kinds of information to query during inference. \citet{lewis_retrieval-augmented_2021}'s RAG follow the same retrieval-based approach but instead integrates a non-parametric seq2seq to improve on tasks outside of open-domain extractive question answering. In the case of multi-task settings, \citet{Maillard2021MultitaskRF} show that it is indeed possible to develop a universal multi-task retriever using non-task-specific methods using a passage and query encoder shared across all tasks.

\end{document}